\documentclass[journa]{IEEEtran}
\usepackage{cite}

\ifCLASSINFOpdf
\else
\fi
\usepackage[cmex10]{amsmath}
\usepackage{graphics}
 \usepackage{graphicx}

\usepackage{amssymb}
\usepackage{amsthm}
\usepackage{amsmath}
\usepackage{threeparttable}
\usepackage{longtable}
\usepackage{color}

\begin{document}
%
\title{A hybrid evolutionary algorithm with importance sampling for multi-dimensional optimization}
%
%
%

\author{Guanghui Huang,~
        Zhifeng Pan
\thanks{Guanghui Huang and Zhifeng Pan are both  with the Department
of Statistics and Accurate Science, College of Mathematics and Statistics, Chongqing University, Chongqing, 401331 China. E-mail:  hgh@cqu.edu.cn(Guanghui Huang), pzf@cqu.edu.cn(Zhifeng Pan).}}
\maketitle

\begin{abstract}
A hybrid evolutionary algorithm with importance sampling method is proposed for multi-dimensional optimization problems  in this paper.
In order to make use of the information provided in the search process, a set of visited solutions is selected to give scores for intervals in each dimension, and they are updated as algorithm proceeds.
Those intervals with higher scores are regarded as good intervals, which are used to estimate the joint distribution of optimal solutions through an interaction between the pool of good genetics, which are the individuals with smaller fitness values.
  And the sampling probabilities for good genetics are determined through an interaction between those estimated good intervals. It is a cross validation mechanism which determines the sampling probabilities for good intervals and genetics,  and the resulted probabilities are used to design crossover, mutation and other stochastic operators with importance sampling method.
        As the selection of genetics and intervals is not directly dependent on the values of fitness, the resulted offsprings may avoid the trap of local optima.
          And a purely random EA is also combined into the proposed algorithm to maintain the diversity of population.
30 benchmark test functions are used to evaluate the performance of the proposed algorithm, and it is found that the proposed hybrid algorithm is an efficient algorithm for multi-dimensional optimization problems considered in this paper.
\end{abstract}


%
\IEEEpeerreviewmaketitle

\section{Introduction}
%
%
%
%

\IEEEPARstart{O}{ptimization} to a problem is a process for seeking  better or best alternative solution from a number of possible solutions \cite{quang2009}.  As the analytical optimal solution is difficult to obtain even for relatively simple application problems, the need for numerical optimization algorithm arises from almost every field of engineering design,
systems operation, decision making, and computer science \cite{yuping2007,jinn2004, zhenguo2004}. In global optimization problems, the particular challenge is that an algorithm may be trapped in the local optima of the objective function when the dimension is high and there are numerous local optima \cite{yw2001}.

Typical conventional search methods include steepest descent methods, conjugate gradient, quadratic programming, and linear approximation methods. These strategies rely on local information of the objective function to decide on their next move in the neighborhood of visited solutions. Their main advantage is the efficiency, however, they tend to be sensitive to starting point selection, and more likely to settle at non-global optima than modern stochastic algorithms \cite{quang2009}.

Modern stochastic algorithms such as evolutionary algorithms (EAs) draw inspiration from biological evolution. They guide the evolution of a set of randomly selected individuals through a number of generations in approaching the global optimum solution, making use of competitive selection, recombination, crossover, mutation or other stochastic operators to generate new solutions \cite{quang2009,zhenguo2004}. They only require information of the objective function itself, and other accessory properties such as differentiability or continuity are not necessary. And EAs essentially work with building blocks, which increase exponentially as the evolution through generations proceeds. This results an efficient exploitation of the given search space.

Modern stochastic optimizers include simulated annealing, Tabu search, genetic algorithms, evolutionary programming, evolution strategies, differential evolution, and others \cite{tuccillo1995,hao2002,friswell1998, krishnakumar1995, sathyanarayan1999, greenwood1995, mulgund1998, seywald1995, hager1994, YaoLiu1999, goldberg1989, kirkpatrick1983, back1991, davis1991, michalewicz1996, auger2006}.  Most of the successful applications of EAs are limited to problems with dimensions below 30 \cite{hager1994, YaoLiu1999, goldberg1989, michalewicz1996}. Only in the last decade, did researchers begin to test their EAs on problems with more than 30 dimensions \cite{Lucidi2002,  Ge1999, Oblow2001, Liang2006, Koumousis2006, Ma2006, Alba2005, Lee2004, Ratnaweera2004, Bergh2004, yw2001}.

To deal with these high dimensional and complex problems effectively and enhance EAs, many researchers  have tried to combine techniques from other research fields into EAs.  The combination of evolutionary algorithm with local search approach is known as \textit{Memetic} or \textit{Hybrid}
algorithm \cite{Garai2013}.  Several new designed hybrid algorithms have been applied to practical problems \cite{Caponio2007, TangYao2007, TangLim2007, Zhu2007, Paenke2009}.  The studies on hybrid algorithm have demonstrated that they converge to high quality solutions more efficiently than their conventional counterparts \cite{quang2009}.  The purpose of this paper is to develop a more efficient hybrid EA for high dimensional optimization problems.

Several local search methods have been successfully combined into EAs.  A robust stochastic genetic algorithm (StGA) for global numerical optimization
is given in \cite{zhenguo2004}, where a stochastic coding scheme based on Gaussian distribution is proposed.
A mutation operator based on Cauchy distribution was proposed as a ``fast evolutionary programming" \cite{YaoLiu1999}, and a further generalization of the mutation operator with L\'{e}vy  distribution was given in \cite{Lee2004}. These algorithms are based on the assumptions about the sampling distributions. In order to avoid the influence of distributional assumption, an non-parameterized importance sampling method is proposed in this paper.

Experimental design methods have been successfully combined into EAs \cite{yuping2007}. Zhang and Leung were the first to combine the orthogonal design into EAs for a discrete optimization problem \cite{KrasnogorSmith2000}, and Li and Smith used Latin squares to improve EAs \cite{OngNair2003}.   Tsai et al. combined the Taguchi method into a genetic algorithm \cite{OngLimZhu2006}. Other researchers set up a marginal model to estimate the distribution of globally optimal solutions for any problem and obtained good results \cite{NomanIba2008,Lozano2004}. On the other hand, the estimation of marginal distribution is not enough for high dimensional optimization problems, due to the number of possible combinations increases exponentially with larger scale of problems.

A relatively simple method is proposed to estimate the joint distribution of optimal solutions in this paper.
It is supposed that the interval which
makes an individual a smaller value of fitness than the value of a similar individual should be given a larger value of probability in the estimated joint distribution, therefore a set of genetics is selected from the visited solutions to give a score for each interval, and those intervals with scores beyond the $75\%$  quantiles are regarded as good intervals for each dimension. On the other hand, the solutions with smaller values of fitness are regarded as good genetics, and those good individuals with more elements falling into good intervals are more likely to be optimal solutions, which should be given a larger probability of selection. At the same time, those good intervals with more good genetics appearing should be given a larger probability of selection. It is a cross validation between good intervals and the pool of good genetics which determines the importance sampling probabilities for good intervals and good genetics in this paper.

Many stochastic algorithms do not memorize places where they have visited, and the information
about the evaluated solutions is not taken into consideration for further search. In order to improve the efficiency of EA, a genetic algorithm that adaptively mutates and never revisits was proposed by \cite{YuenChow2009}. And an evolutionary algorithm based on the entire previous search history (HdEA) was proposed in \cite{ChowYuen2011}. However,  there are more and more visited solutions needed to be memorized as algorithm proceeds, such that the requirement of memory may be extremely large. In order to use the information provided by the previous search process, and to avoid the extra requirement of memory ability,  only part of the visited solutions are selected and used to give scores for the intervals in this paper.  They are updated from one generation to the next, and the requirement of memory is a parameter which can be adjusted during the process of algorithm design.

Premature population convergence about a local optimum is a common problem of traditional genetic algorithms \cite{holland1992}. It is a result of individuals hastily congregating within a small region of the search space \cite{ginleymaher2011}. Maintaining a diverse population is very important for evolutionary algorithms, which means that the selection of individuals can not only dependent on their fitness scores, and other principle such as the diversity proposed in \cite{ginleymaher2011} should be taken into consideration. The distributions of importance sampling for individuals and intervals are  determined  through a cross validation mechanism between the pool of good genetics and the good intervals in this paper,  which is not related to the values of fitness directly.  And a purely random EA is combined into the proposed algorithm to maintain the diversity of individuals in this paper.

$30$ test functions and $9$ benchmark evolutionary algorithms are selected to evaluate the performance of the proposed algorithm.
There are $6$ new optimal solutions found in our numerical investigations,
 $10$ solutions similar to the best results reported in the literature, and $8$ solutions closed to the best results. On the other hand,  there are $6$ test functions where the proposed algorithm can not find the optimal solutions efficiently. However, the proposed algorithm has the smallest number of fitness values which are different from the optimal solutions with respect to the order of magnitude among the algorithms considered in this paper.

The remainder of this paper is structured as follows. Section \ref{sec-1} describes the problem of optimization for multidimensional functions. The details of hybrid EA are given in Section \ref{sec-2}.
Section \ref{sec-3} is devoted to the empirical investigations of the proposed algorithm through 30 test functions. And conclusions and discussions  are given in Section \ref{sec-4}.

\section{Optimization problem}\label{sec-1}

The problem we consider is an unconstrained global optimization problem
\begin{equation}
\min_{x\in G} f(x),
\end{equation}
where $x=(x_1,x_2,\cdots,x_n)\in\mathrm{R^n}$ is a vector with $n$ elements,  $G=\left\{ x\in \mathrm{R^n} | b_i^l \le x_i \le b_i^u, i=1,2,\cdots,n \right\}$ is a subset of $R^n$, where $b_i^l$ and $b_i^u$ are the lower and upper boundaries of $x_i$ respectively. The value of objective function at point $x$ is called the fitness value of $x$ in this paper. The purpose of optimization is to find the solutions which make the objective function reach its minimum value.

\section{Hybrid EA with importance sampling}\label{sec-2}

\begin{figure}
  \centering
    \caption{HisEA for multi-dimensional optimization problems.}\label{flowchart}
    \vskip 1em
  \includegraphics[width=0.95 \linewidth]{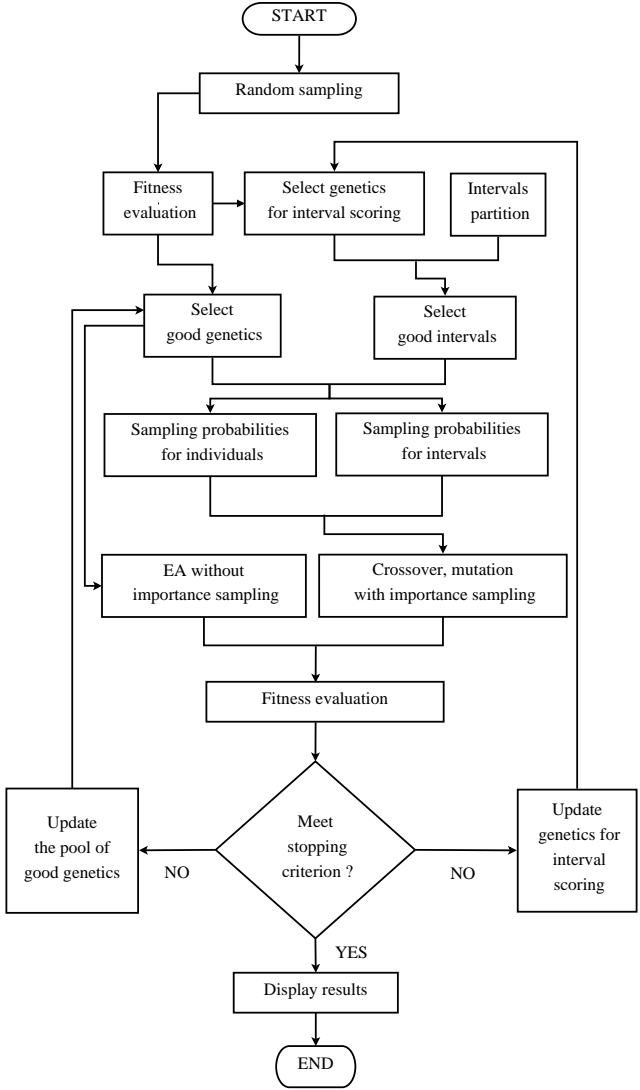}
\end{figure}

Canonical EA is an optimization algorithm based on population, where individuals are used to generate the offspring generation with genetic operators, such as mutation, crossover, and selection.
The individuals with smaller values of fitness are survival from the evolution of population. While the information provided by those individuals which are not survival is completely dropped in further searching process. Some researchers have suggested to use those information efficiently to improve the performance of EAs \cite{YuenChow2009, ChowYuen2011}. Following this line, the information obtained in the process of searching is used to design new crossover operator,  mutation operator, interpolation operator with importance sampling method in this paper.

 \subsection{Initiation}
 The individuals of first generation are randomly generated within the search space, where the size of the first generation $N_f$ is a predetermined parameter. There are $N_g$ individuals chosen to be the pool of good genetics.  The range of search in each dimension is partitioned into $N_p$ subintervals with equal length.   And $N_s$ is the base number of new generated individuals,  where the numbers of new generated solutions for crossover, mutation and interpolation operators are several times of $N_s$ respectively in the following sections.  There are only four parameters needed to be determined before the application of the hybrid EA with importance sampling method (HisEA).  Figure \ref{flowchart} is the flow chart of HisEA.

\subsection{Fitness scores of individuals}

Suppose the individuals in the current pool of good genetics are $x_1,x_2,\cdots,x_{N_g}$, whose values of fitness are $f_1 \le f_2 \le \cdots \le f_{N_g}$ with increasing order, and the maximum value of fitness in the current search history is denoted as
$f_{max}$.  The score for the ith individual is defined as
\begin{equation}
w_i = \frac{f_{max}-f_{i}}{{N_g} f_{max}-\sum_{i=1}^{{N_g}}{f_i}},
\end{equation}
which indicates that the individual with smaller value of fitness will be given a relatively larger score among the current pool of good genetics. And $f_{max}$ will be updated in the following search process, such that the score for each individual is changeable to update the new information achieved in  the process of search.

\subsection{Scores of intervals}

As each dimension of the search space has been partitioned into $N_p$ equal subintervals, the length of one interval
for the ith dimension is
\begin{equation}
|\delta_i^k|=\frac{b_i^u-b_i^l}{N_p},
\end{equation}
where $\delta_i^k=[\Delta_i^k,\Delta_{i}^{k+1})$ is the kth interval of the ith dimension, and $b_i^l=\Delta_i^{1} < \Delta_i^{2} < \cdots < \Delta_i^{N_p+1}=b_i^u$ are the partition points of this dimension, and
\begin{equation}
\Delta_i^{j+1}=\Delta_i^{j} + |\delta_i^k|.
\end{equation}

\subsubsection{Selection of scoring genetics for intervals}
A pool of genetics $G_{s}$ is selected from all of the evaluated solutions to give a score for each interval of every dimension with the following algorithm.
\begin{enumerate}
\item[(1)]   Initiation $G_{s}$ according to the first dimension. Denote the genetics in the first generation as  $x_j=(x_{j1},x_{j2},\cdots,x_{jn})$, $j=1,2,\cdots,N_f$. For the kth interval of the first dimension $\delta_{1}^{k}$,   if $x_{j_m 1} \in \delta_{1}^{k}$, $m=1,2,\cdots,k_m$, where
    $f(x_{j_1}) \le f(x_{j_2}) \le \cdots \le f(x_{j_{k_m}})$,  then $x_{j_1}$ and $x_{j_2}$ are put into $G_s$. In other words, the first two solutions whose elements of the first dimension are in the same interval are selected according to their fitness. And the solution with maximum value of fitness is included in $G_s$.
\item[(2)] Repeat step (1) for $k=1,2,\cdots,N_p$.
\item[(3)] Update $G_s$ according to new evaluated solutions.  If a new evaluated solution $x$ is belong to the kth interval, that is to say $x_1 \in \delta_1^k$, the first two solutions with smaller values of fitness among the previous selected genetics and the new genetic are selected for the kth interval. And the solution with maximum value of fitness is updated.
\end{enumerate}

\subsubsection{Selection of good intervals}
$G_s$ is used to give a score for each subinterval of every dimension, and denote $S(i,j)$ the score for the jth interval of the ith dimension, $i=1,2,\cdots,n$, $j=1,2,\cdots,N_p$. The score matrix $S(i,j)$ is used to determine the  good intervals for each dimension with the following algorithm.
\begin{enumerate}
\item[(1)]  Initiation. Set $S(i,j)=0$, $i=1,2,\cdots,n$, $j=1,2,\cdots,N_p$.
\item[(2)]  For the ith dimension and the jth interval, find the genetics in $G_s$ whose ith elements are
            in the jth subinterval $\delta_i^{j}$, denote as $x_1, x_2, \cdots, x_{N_{in}}$, where  $N_{in}$ is the number of genetics appearing in the jth subinterval.
            \begin{enumerate}
            \item[(i)]   Case 1. $N_{in} < 2$, set $S(i,j)=S(i,j)$.
            \item[(ii)]  Case 2. $N_{in} \ge 2$, select the first two genetics  $x_1$ and $x_2$ according to
                          their order in $G_s$, and denote their weights as $w_1$ and $w_2$ ($w_2<w_1$). For $k=1,2,\cdots,n$, denote $x_{1k} \in \delta_{k}^{m_1}$ and $x_{2k} \in \delta_{k}^{m_2}$, set
            \begin{align}
            S(i,k)=\left\{
                      \begin{aligned}
                       S(i,k)+w_1+w_2,       & \text{if~}    m_1=m_2;   \\
                       S(i,k)+2 \times w_1,  & \text{if~}    m_1 \ne m_2.
                     \end{aligned}
                      \right.
            \end{align}
            \end{enumerate}
\item[(3)]  Repeat step (2) for  $j=1,2,\cdots,N_p$  and $i=1,2,\cdots,n$.
\item[(4)] Suppose $H(k)$ is the $75\%$ quantile of the kth row of the score matrix $S$. If $S(k,m) \ge H(k)$, the subinterval $\delta_{k}^m$ is said to be a good interval for the kth dimension.
\item[(5)] If $\delta_k^{m}$ and  $\delta_k^{m+1}$  are both good intervals, set $\tilde{\delta}_k^m=[\tilde{\Delta}_{k}^m,\tilde{\Delta}_{k}^{m+2}]= \delta_k^{m} \bigcup \delta_k^{m+1}$.
\item[(6)] Repeat Step (5) until there is no more subinterval to be combined. And $\tilde{\delta}_k^m$ is
           said to be the mth good interval for the kth dimension.
\end{enumerate}

\subsection{Sampling probabilities of individuals}

The individuals in the pool of good genetics are chosen according to their values of fitness. In order to describe the distributional information among all of the dimensions, the sampling probabilities of individuals are chosen in the following way.  Denote  $I_i=(I_{i1},I_{i2},\cdots,I_{in})$  the indicator function for $x_i=(x_{i1},x_{i2},\cdots,x_{in})$.  Let
\begin{equation}
I_{ik}=\left\{
\begin{array}{ll}
1, & \text{if } x_{ik} \in \tilde{\delta}_{k}^{m}, \text{which is a good interval}  ; \\
0, & \text{otherwise}.
\end{array}
 \right.
\end{equation}
The score of $x_i$ is
\begin{equation}
c_i=\sum_{k=1}^{n} I_{ik},
\end{equation}
which is the number of elements falling into the good intervals.

Denote $p_i$ the probability that the ith individual is chosen among the $N_g$ individuals in the pool of good genetics,
\begin{equation}
p_i = \frac{c_i}{ \sum_{j=1}^{N_g} c_j},
\end{equation}
which means that it is more possible to be chosen for those individuals with more elements falling into the good intervals.

The sampling probabilities for individuals are not directly based on the values of fitness in this paper, which can be regarded  as an alternative choice to maintain the diversity of population.

\subsection{Sampling probabilities of intervals}

There is a cross validation mechanism between the chosen good intervals and the individuals in the pool of good genetics, which is used to determine the sampling probabilities of the individuals in the previous section, and to determine the sampling probabilities of intervals in this section with the following algorithm.
\begin{enumerate}
\item[(1)] Initiation. Let $q(m,k)=0$ be the number of individuals falling into the kth good interval of the mth dimension, $m=1,2,\cdots,n$, and $k=1,2,\cdots,N_{i}^{m}$, where $N_{i}^{m}$ is the number of good intervals of the mth dimension.
\item[(2)] Denote $x_i = (x_{i1},x_{i2},\cdots,x_{in})$ an individual in the pool of good genetics, if $x_{im} \in  \tilde{\delta}_{m}^{k}$, where $k$ is some integer between 1 and $N_{i}^{m}$, let
     $q(m,k)=q(m,k)+1$, otherwise $q(m,k)=q(m,k)$.
\item[(3)] Repeat step (2) for $m=1,2,\cdots,n$.
\item[(4)] Repeat step (2) and (3) for $i=1,2,\cdots,N_g$.
\item[(5)] The sampling probability for the kth good interval of the mth dimension is
\begin{equation}
p(m,k)=\frac{q(m,k)}{\sum_{k=1}^{N_{i}^{m}} q(m,k) }.
\end{equation}
\end{enumerate}

The estimated sampling probabilities for individuals and intervals are used to design a crossover operator, two kinds of mutation operators,  and an interpolation operator  with importance sampling method in the following sections.

\subsection{Crossover operator with importance sampling}

Crossover operator is used to generate new individuals from their parents.
As the elements in the good intervals are more likely to be the optimal solutions, they will be kept in the offsprings, and those elements not in the good intervals are replaced by the elements of the other parent which are in the good intervals as the following algorithm.
\begin{enumerate}
\item[(1)]  Sampling two different individuals from the pool of good genetics with the importance sampling probabilities $p_i,i=1,2,\cdots,N_g$, say $x_1$ and $x_2$.
\item[(2)] Find the elements in the good intervals for  $x_1$ and $x_2$, whose positions are indicated by two indicators, denoted as $I_{1}=(I_{11},I_{12},\cdots,I_{1n})$ and $I_{2}=(I_{21},I_{22},\cdots,I_{2n})$ respectively, where $I_{ij}=1$ means that the jth element in the ith individual is falling into the good intervals, $i=1,2$, $j=1,2,\cdots,n$. Otherwise $I_{ij}=0$.
\item[(3)] Generate one individual $y$ from $x_1$ with elements chosen from $x_2$ by the following algorithm:
    \begin{enumerate}
    \item[(i)] If $I_{1j}=0$ and $I_{2j}=0$, $y_{j}=x_{1j}$;
    \item[(ii)] If $I_{1j}=1$ and $I_{2j}=0$, $y_{j}=x_{1j}$;
    \item[(iii)] If $I_{1j}=0$ and $I_{2j}=1$, $y_{j}=x_{2j}$;
    \item[(iv)] If $I_{1j}=1$ and $I_{2j}=1$, $y_{j}=x_{2j}$.
    \end{enumerate}
\item[(4)] Repeat step (3) for $x_2$.
\item[(5)] Repeat step (1) to (4) $N_{s}$ times to generate a set of new genetics.
\end{enumerate}

The proposed algorithm is based on the pool of good genetics, which are chosen according to their  values of fitness.
 On the other hand, the two parents to generate new individuals are sampled with the importance sampling probabilities, which are not directly related to the values of fitness. And the result of crossover is related to the estimation of good intervals, which can be regarded as the estimation of the joint distribution of the optimal solutions. This is the difference between the proposed hybrid algorithm and the traditional EAs.

\subsection{Mutation operators with importance sampling}

There are two kinds of mutation operators proposed in this section, which are all based on the importance sampling probabilities.

\subsubsection{Locally adjusting algorithm}

There may be some individuals in the pool of good genetics whose elements are not all falling into the good intervals. In order to make those individuals look more like good genetics, a locally adjusting algorithm is proposed as the following steps.
\begin{enumerate}
\item[(1)] Select one of the individuals in the pool of good genetics according to the probabilities $p_i, i=1,2,\cdots,N_g$, denote as $x=(x_1,x_2,\cdots,x_n)$, and denote $y$ the individual to be generated, set $y=x$.
\item[(2)] Mutation for the kth dimension.
 If there dose not exist  any $m$ such that $x_k \in \tilde{\delta}_{k}^{m}$, select one of the good intervals of the kth dimension according to $q(k,m), m=1,2,\cdots,N_k$, denoted as $[\tilde{\Delta}_{k}^{m},\tilde{\Delta}_{k}^{m+1}]$, and $y_k$ is adjusted as  $y_k=\tilde{\Delta}_{k}^{m}+(\tilde{\Delta}_{k}^{m+1}-\tilde{\Delta}_{k}^{m})*U$, where $U$ is an uniformly distributed random variable on $[0,1]$.
\item[(3)] Repeat step 2 for $k=1,2,\cdots,n$. If there is no dimension to be adjusted, there is no new genetic to be generated in this run.
\item[(4)] Repeat $N_s$   times to generate a set of new individuals.
\end{enumerate}

\subsubsection{Entirely adjusting algorithm}

Another mutation algorithm is proposed to explore the visited space as the following steps.

\begin{enumerate}
\item[(1)] Select one of individuals in the pool of good genetics according to the probabilities $p_i, i=1,2,\cdots,N_g$, denote as $x=(x_1,x_2,\cdots,x_n)$, and denote $y$ the individual to be generated.
\item[(2)] Mutation for the kth dimension with the following algorithm:
        \begin{enumerate}
          \item[(i)] If there dose not exist  any $m$ such that $x_k \in \tilde{\delta}_{k}^{m}$, select one of the good intervals with  $q(k,m), m=1,2,\cdots,N_k$, denoted as $[\tilde{\Delta}_{k}^{m},\tilde{\Delta}_{k}^{m+1}]$, and $y_k$ is adjusted as  $y_k=\tilde{\Delta}_{k}^{m}+(\tilde{\Delta}_{k}^{m+1}-\tilde{\Delta}_{k}^{m})*U$.
          \item[(ii)] If there exists some $m$, such that $x_k \in [\tilde{\Delta}_{k}^{m},\tilde{\Delta}_{k}^{m+1}]$, $y_k$ is adjusted as  $y_k=\tilde{\Delta}_{k}^{m}+(\tilde{\Delta}_{k}^{m+1}-\tilde{\Delta}_{k}^{m})*U$, where $U$ is an uniformly distributed random variable on $[0,1]$.
        \end{enumerate}
\item[(3)] Repeat step (2) for $k=1,2,\cdots,n$.
\item[(4)] Repeat step (1) to (3) $N_s \times 2$  times to generate a set of new individuals.
\end{enumerate}

The difference between these two kinds of mutation operators is that
the elements falling into the good intervals are not adjusted by the locally adjusting algorithm, while which are adjusted by the entirely adjusted algorithm.

\subsection{Interpolation operator with importance sampling}

In order to search the space between two suboptimal solutions, an interpolation operator is adopted in this paper, where
the estimated good intervals are used to guide the direction of search as the following steps.
\begin{enumerate}
\item[(1)] Randomly choose two individuals in the pool of good genetics according to the probabilities $p_i$,$ i=1$, $2$, $\cdots$, $N_g$, denoted as $x_1$ and $x_2$.
\item[(2)] Generate the element for the ith dimension with the following algorithm:
  \begin{enumerate}
  \item[(i)]
  If there exists two good intervals $\tilde{\delta}_i^{k}$ and $\tilde{\delta}_i^{m}$ such that
                  $x_{1i} \in \tilde{\delta}_i^{k}$ and $x_{2i} \in \tilde{\delta}_i^{m}$. Set $l=[\frac{k+m}{2}]$, where $[x]$ denotes the largest integer which is less than or equal to $x$, and generate the ith dimension for the new individual $y$ as
                  \begin{equation}
                  y_i = \tilde{\Delta}_{i}^{l} + (\tilde{\Delta}_{i}^{l+1}-\tilde{\Delta}_{i}^{l})\times U,
                  \end{equation}
                  where $U$ is uniformly distributed on $[0,1]$.
  \item[(ii)] If there exists one good interval $\tilde{\delta}_{i}^{k}$ such that $x_{1i} \in \tilde{\delta}_{i}^{k}$, and no good interval to contain $x_{2i}$. The ith dimension for the new individual $y$ is
      \begin{equation}
      y_i = \tilde{\Delta}_{i}^{k} + (\tilde{\Delta}_{i}^{k+1}-\tilde{\Delta}_{i}^{k}) \times U.
      \end{equation}
  \item[(iii)]  If there exists one good interval $\tilde{\delta}_{i}^{m}$ such that $x_{2i} \in \tilde{\delta}_{i}^{m}$, and no good interval to contain $x_{1i}$. The ith dimension for the new individual $y$ is
      \begin{equation}
      y_i = \tilde{\Delta}_{i}^{m} + (\tilde{\Delta}_{i}^{m+1}-\tilde{\Delta}_{i}^{m}) \times U.
      \end{equation}
 \item[(iv)] If there exists no good interval to contain any  of the two samples $x_{1i}$ and $x_{2i}$, a good interval for the ith dimension is randomly selected according to the probabilities $q(i,k)$, $k=1,2,\cdots, N_i$, where $N_i$ is the number of good intervals for the ith dimension, denoted as $\tilde{\delta}_{i}^{m}$. The ith dimension for the new individual is
     \begin{equation}
     y_i = \tilde{\Delta}_{i}^{m} + (\tilde{\Delta}_{i}^{m+1}-\tilde{\Delta}_{i}^{m}) \times U.
     \end{equation}
  \end{enumerate}
\item[(3)] Repeat Step 2 for $k=1,2,\cdots,n$ to generate a new individual.
\item[(4)] Repeat Step 1 to Step 3  $N_s$ times to generate a set of new individuals.
\end{enumerate}

\subsection{Random sampling}

In order to explore the search space,
there are two kinds of random sampling methods adopted in this paper, one of which is based on the probabilities of importance sampling, and the other one is not related to the information obtained in the process of searching.

\subsubsection{Importance sampling algorithm}
Importance sampling algorithm is designed to explore the search space, where the estimated distribution of optimal solutions is involved in the following steps.
\begin{enumerate}
\item[(1)] For the ith dimension, one of the estimated good subinterval is sampled according to the probabilities $q(i,k),k=1,2,\cdots,N_{i}$, denoted  as $[\tilde{\Delta}_{i}^{start},\tilde{\Delta}_{i}^{end}]$.
\item[(2)] Randomly sampling one sample from $[\tilde{\Delta}_{i}^{start},\tilde{\Delta}_{i}^{end}]$ as
\begin{equation}
y_i=\tilde{\Delta}_{i}^{start}+(\tilde{\Delta}_{i}^{end}-\tilde{\Delta}_{i}^{start})\times U,
\end{equation}
where $U$ is uniformly distributed within $[0,1]$.
\item[(3)] Repeat Step 1 to Step 2 for $i=1,2,\cdots,n$.
\item[(4)] Repeat Step 1 to Step 3 $N_{s}$ times to generate a set of new individuals.
\end{enumerate}

As more and more individuals are generated from the estimated good intervals,   the resolution of these intervals is improved.

\subsubsection{Purely random sampling}
In order to keep the diversity of the chosen good genetics, and to reduce the risk of premature,  a purely random sampling method is adopted as the following steps.
\begin{enumerate}
\item[(1)] For the ith dimension, $y_i$ of individual $y$ is
 \begin{equation}
 y_i = b_{i}^{l} + (b_{i}^{u}-b_{i}^{l}) \times U,
 \end{equation}
 where $b_{i}^{l}$  and $b_{i}^{u}$ are the lower  and upper boundaries for the ith dimension respectively.
 \item[(2)] Repeat Step 1 for $i=1,2,\cdots,n$.
 \item[(3)] Repeat Step 1 to Step 2  $N_s \times 5$ times to generate a set of new individuals.
\end{enumerate}

\subsection{Purely random EA}

In order to keep the diversity of genetics, and to escape the trap of local optimal solutions, an evolutionary algorithm with purely random crossover and mutation operators is adopted in this paper, which is dependent on the pool of good genetics, but  does not use the information from the previous search process.

\subsubsection{Purely random crossover}
 A purely random crossover operator is adopted in this paper as following steps.
\begin{enumerate}
\item[(1)] Select two individuals in the pool of good genetics with equally possibility, denoted as $x_1$ and $x_2$.
\item[(2)] Denote $y_1$ and $y_2$ the new individuals  to be generated.
\item[(3)] For the ith dimension, randomly sample a number $E$, where $E$ is a binomial distributed variable $E \sim B(1,1/2)$. The elements of $y_1$ and $y_2$ are determined with the following algorithm.
    \begin{enumerate}
    \item[(i)] If $E=1$, $y_{1i}=x_{2i}$, and $y_{2i}=x_{1i}$.
    \item[(ii)] If $E=0$, $y_{1i}=x_{1i}$, and $y_{2i}=x_{2i}$.
    \end{enumerate}
\item[(4)] Repeat Step 3 for $i=1,2,\cdots,n$.
\item[(5)] Repeat Step 1 to Step 4 $ N_s \times 5$ times to generate a set of new individuals.
\end{enumerate}

As the result of random trail $E$ is equally distributed between $0$ and $1$, the crossover between $x_1$ and $x_2$ is purely random, which is designed to maintain the diversity of population  in this paper.

\subsubsection{Purely random mutation}

A similar algorithm for mutation is adopted in this paper, where the element of the solution is randomly selected to be mutated with the following algorithm.
\begin{enumerate}
\item[(1)] Randomly select one individual in the pool of good genetics with equally probabilities, denoted as $x$.
\item[(2)] Randomly sample a value of $E \sim B(1,1/2)$. Denote the new genetic as $y$, whose element in the ith dimension is determined by the following algorithm.
\begin{enumerate}
\item[(i)] If $E=1$, $y_i=\Delta_{i}^{k}+(\Delta_{i}^{k+1}-\Delta_{i}^{k}) \times U$, where $x_i \in \delta_{i}^{k}$.
\item[(i)] If $E=0$, $y_i = x_{i}$.
\end{enumerate}
\item[(3)] Repeat Step 2 for $i=1,2,\cdots,n$.
\item[(4)] Repeat Step 1 to Step 3 $N_s\times 10$ times to generate a set of individuals.
\end{enumerate}

The total number of new generated individuals with all of the previous operators is $N_s \times 32$ for each run of the hybrid algorithm, where individuals generated without the information obtained in the process of search are $N_s \times 5$, which is designed to maintain the diversity of population.

\subsection{Mature condition}

A pool of good genetics is used to generate new individuals, whose values of fitness are evaluated  and compared to their parents, and a new pool of good genetics is selected from the parents and offsprings according to their values of fitness. Denote $X_{n \times N_g}$  the former pool of good genetics, and $Y_{n \times N_g}$  the new pool of good genetics.  The stopping condition is based on the result of comparison between $X$ and $Y$ with the following algorithm.

Select a set of quantiles, denoted as $0<\alpha_1<\alpha_2<\cdots<\alpha_m<1$, where $m$ is the number of quantiles to be taken into consideration. The quantiles of $X$ for each dimension are denoted as
\begin{equation}
\tilde{X}=\left(
\begin{array}{cccc}
x_{1}^{\alpha_1} & x_{1}^{\alpha_2} & \cdots & x_{1}^{\alpha_m} \\
x_{2}^{\alpha_1} & x_{2}^{\alpha_2} & \cdots & x_{2}^{\alpha_m} \\
\vdots & \vdots & \vdots & \vdots \\
x_{n}^{\alpha_1} & x_{n}^{\alpha_2} & \cdots & x_{n}^{\alpha_m}
\end{array}
\right),
\end{equation}
where $x_{i}^{\alpha_{k}}$ is the kth quantile for the ith dimension, $i=1,2,\cdots,n$, and $k=1,2,\cdots,m$. The similar quantiles for $Y$ is denoted as $\tilde{Y}$. The difference between those two kinds of quantiles is
\begin{equation}
\parallel \tilde{X} - \tilde{Y}  \parallel = \max_{i} { \max_{k}  { \mid x_{i}^{\alpha_k} - y_{i}^{\alpha_k} \mid}},
\end{equation}
and the hybrid EA  is stopped when $\parallel \tilde{X} - \tilde{Y}  \parallel \le 1.0\mathrm{E}-9$ or the number of loops is beyond $300$ times, where the quantiles are those points from $0.05$ to $0.95$ with step length $0.05$  in this paper.

%
%
%
%
\section{Empirical investigations}\label{sec-3}

To evaluate the performance of the proposed algorithm, the optimal values of fitness founded by HisEA are compare to their counterparts of 9 benchmark evolutionary algorithms for 30  test functions in this paper.

\subsection{Algorithms for comparison}

\subsubsection{HdEA}
HdEA  is an evolutionary algorithm that uses the entire search
history to improve its mutation strategy \cite{ChowYuen2011}. It uses the fitness
function approximated from the search history to perform
mutation. Since the proposed mutation operator is adaptive
and parameter-less, HdEA has only three control parameters:
neighborhood size, population size, and crossover rate.
The source code of HdEA is available
at http://www.ee.cityu.edu.hk/¡«syyuen/Public/Code.html.

\subsubsection{RCGA-UNDX}
Real Coded GA With Uni-Modal Normal
Distribution Crossover (RCGA-UNDX) is a real coded GA that deals with continuous search
spaces \cite{OnoKobayashi2007,ChowYuen2011}. It applies the uni-modal normal distribution crossover
(UNDX) to preserve the statistics of the population. UNDX
is a multiparent genetic operator in which the distribution
of the corresponding offspring follows the distribution of the
parents.

\subsubsection{CMA-ES}
Covariance Matrix Adaptation Evolution
Strategy (CMA-ES)  is an evolution
strategy that adapts the full covariance matrix of a normal
search (mutation) distribution \cite{Hansen2007,ChowYuen2011}. An important property of CMA-ES
is its invariance against linear transformations of the search
space. The underlying idea is to gather information about
successful search steps to modify the covariance matrix of
the mutation distribution in a de-randomized, goal directed
fashion. Changes to the covariance matrix are such that
variances in directions of the search space that have previously
been successful are increased, while those in other directions
decrease passively. The accumulation of information over a
number of search steps makes it possible to reliably adapt the
covariance matrix even when using small populations. CMA-ES
is designed with the emphasis that the same parameters
are used in all applications in order to be ``parameter-less.''
The source code of CMA-ES is taken from \cite{Hansen2007} (Aug. 2007
version).

\subsubsection{DE}

Differential evolution (DE )
is a stochastic search algorithm \cite{StornPrice1997,ChowYuen2011}. The basic idea behind DE
is a scheme that generates trial parameter vectors. DE adds
the weighted difference between two population vectors to a
mutant vector, and the trial vector is the crossover between
the mutant vector and the parent vector. By doing so, no
separate probability distribution is used, which makes the
scheme completely self-organizing.

\subsubsection{ODE}
 Opposition-based differential evolution (ODE)
utilizes the concept of opposition-based learning (OBL) \cite{Tizhoosh2005}
to accelerate the convergence rate of DE. The main idea
behind OBL is the simultaneous considerations of a solution
and its corresponding opposite solution. ODE considers the
evaluations of the opposite solution in a generation depending
on a jumping rate \cite{RahnamayanTizhooshSalama2008,Tizhoosh2005,ChowYuen2011}.

\subsubsection{DEahcSPX}
Differential Evolution With Adaptive
Hill-Climbing Simplex Crossover (DEahcSPX)
  attempts to accelerate the classic DE by a local search
strategy, named adaptive hill-climbing crossover-based local
search. It adopts the simplex crossover operation (SPX)
to generate offspring individual for hill-climbing \cite{ChowYuen2011,NomanIba2008,Tizhoosh2005}.

\subsubsection{DPSO}
Dissipative Particle Swarm Optimization
(DPSO) is a modified PSO which introduces
random mutation that helps particles to escape from local
minima. Its formula is described as follows:
If $\eta_3 < C_V$  then $V_i = \eta_4 \times V_{max}/C_m$
where $\eta_3$  and  $\eta_4$   are uniformly distributed random variables in
the range $[0, 1]$, $C_V$ is the mutation rate to control the velocity,
$C_m$  is a constant to control the extent of mutation, and  $V_{max}$
is the maximum velocity \cite{XieZhangYang2002, ChowYuen2011} .

\subsubsection{SEPSO}
PSO With Spatial Particle Extension
(SEPSO) is another modified PSO which introduces
the spatial particle extension model to increase the
diversity. When particles start to cluster and collide, they
bounce off by adjusting their velocities \cite{ChowYuen2011, KrinkVesterstromRiget2002}.

\subsubsection{EDA}
EDA is based on
undirected graphical model and Bayesian network. The source
code of the EDA is taken from \cite{SantanaEchegoyen2009} (Feb. 2009 version).
The implementation is conceived to allow the user different
combinations of selection, learning, sampling, and local search
procedures \cite{SantanaEchegoyen2009, ChowYuen2011}.

Each of the above algorithms was executed to some of the test functions, and the results were reported in \cite{ChowYuen2011} and the references  therein. We use existing results for a direct comparison in this section.

\subsection{Simulations and results}

\begin{figure}
 \caption{Convergence of HisEA for 2-dimensional test functions.}\label{lowerd}
  \includegraphics[width=0.95\linewidth]{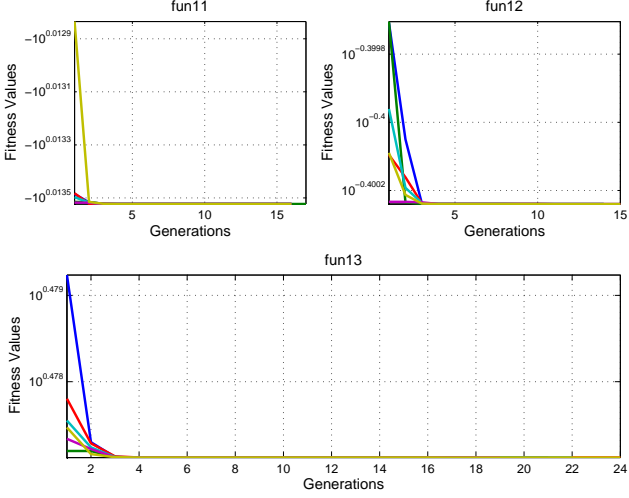}
\end{figure}

\begin{figure}
    \caption{Convergence of HisEA for 30-dimensional test functions.}\label{higher}
  \includegraphics[width=0.95\linewidth]{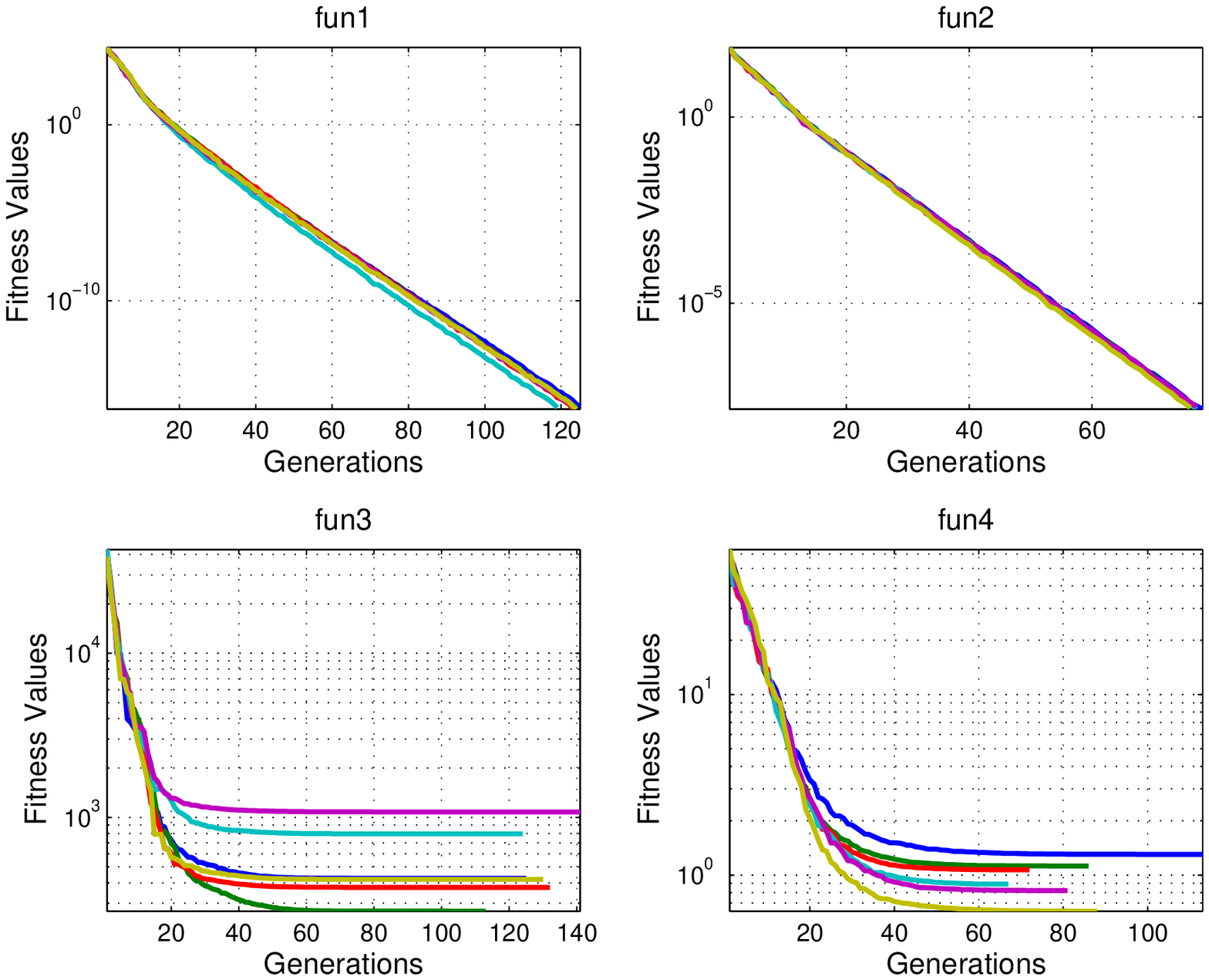}
    \end{figure}
    \begin{figure}
  \includegraphics[width=0.95\linewidth]{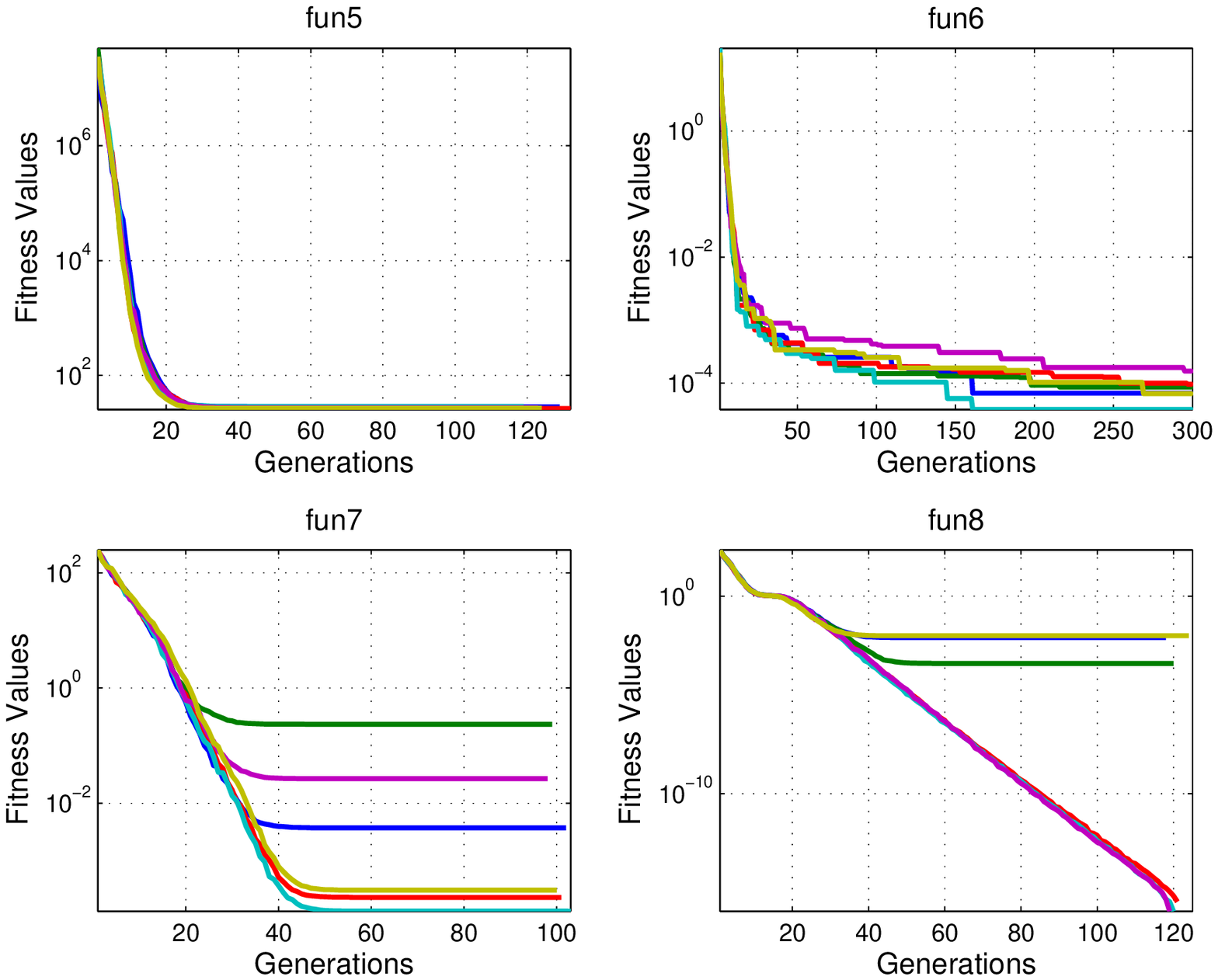}
      \end{figure}
    \begin{figure}
  \includegraphics[width=0.95\linewidth]{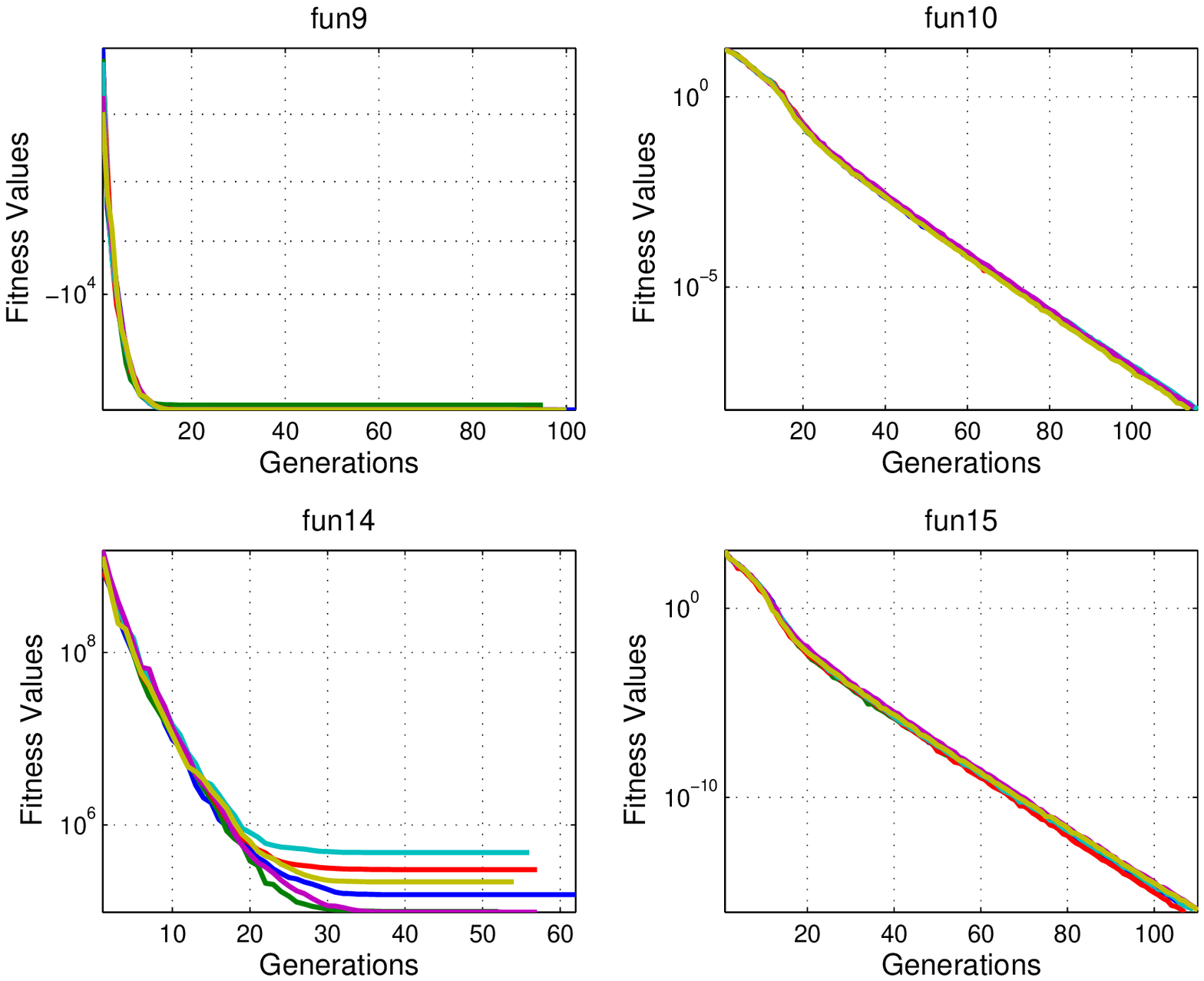}
      \end{figure}
    \begin{figure}
    \includegraphics[width=0.95\linewidth]{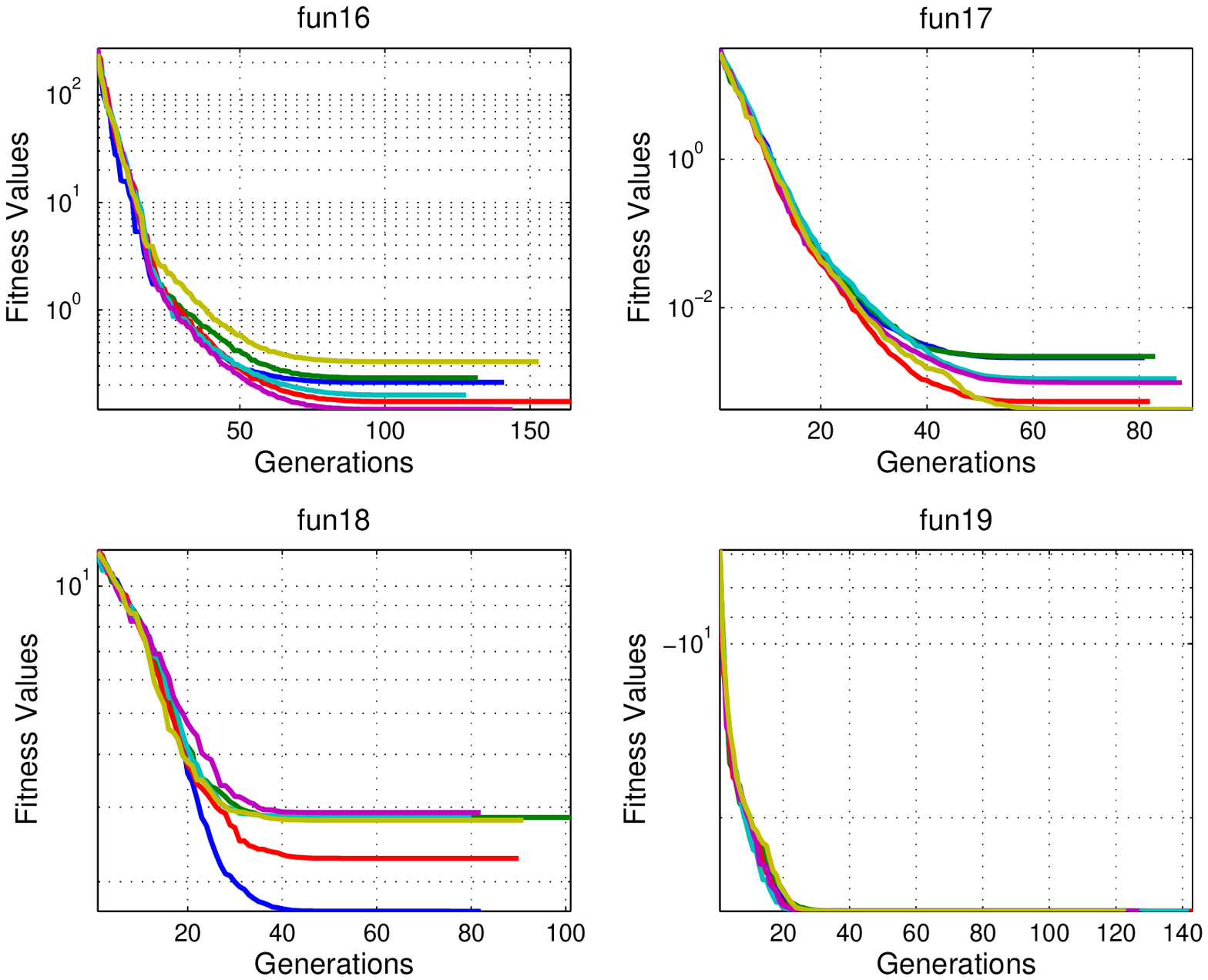}
        \end{figure}
    \begin{figure}
      \includegraphics[width=0.95\linewidth]{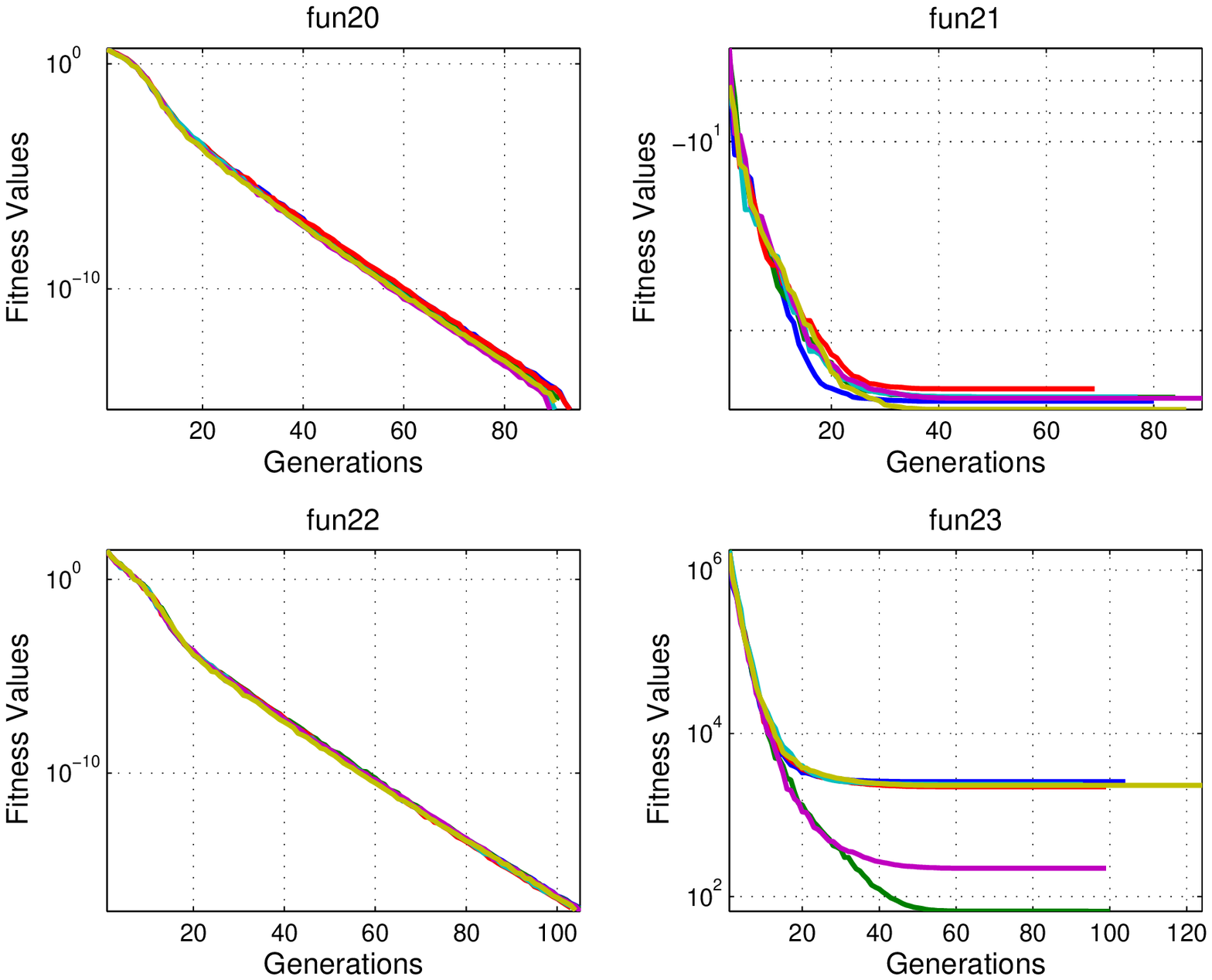}
          \end{figure}
    \begin{figure}
        \includegraphics[width=0.95\linewidth]{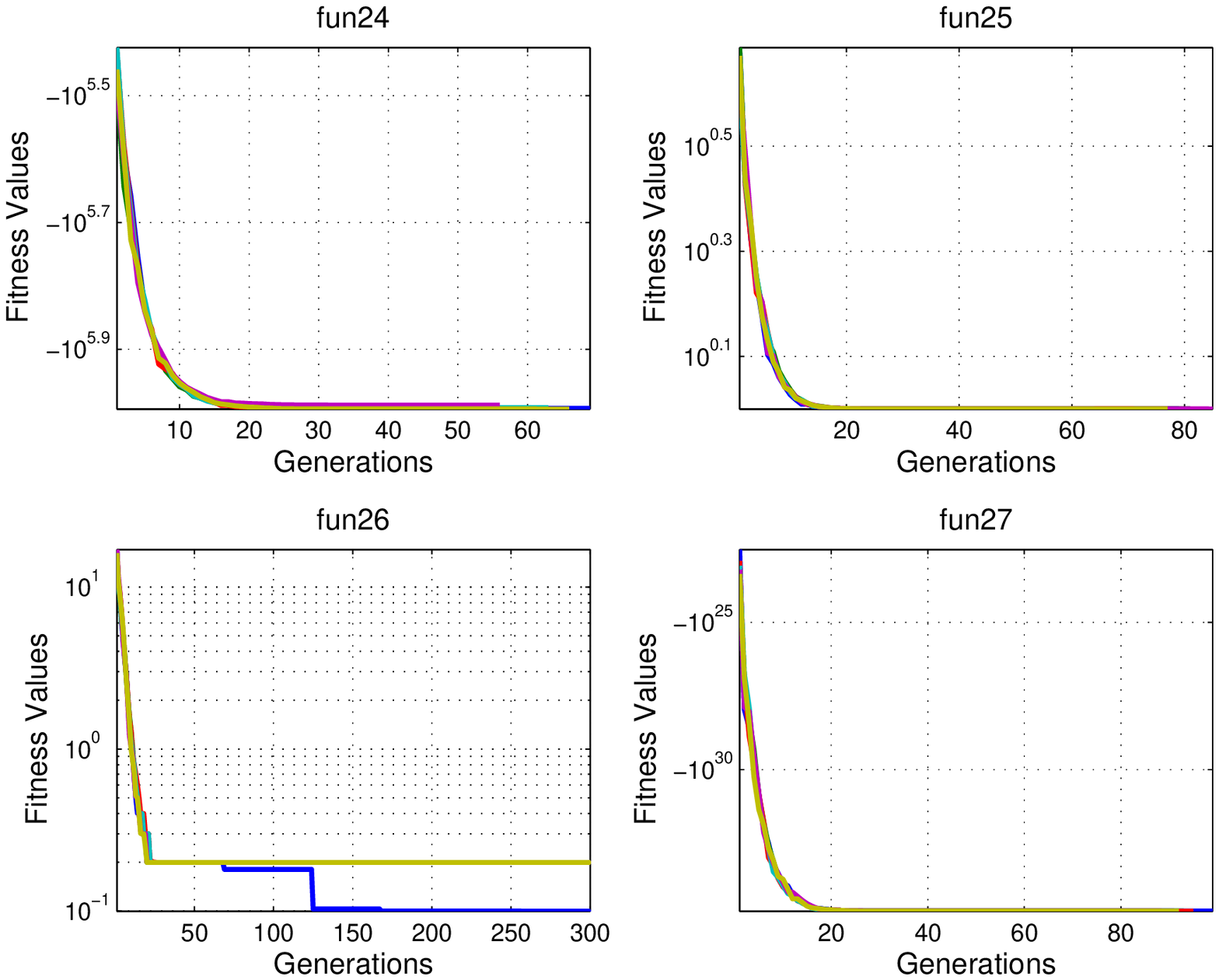}
  \end{figure}
     \begin{figure}
        \includegraphics[width=0.95\linewidth]{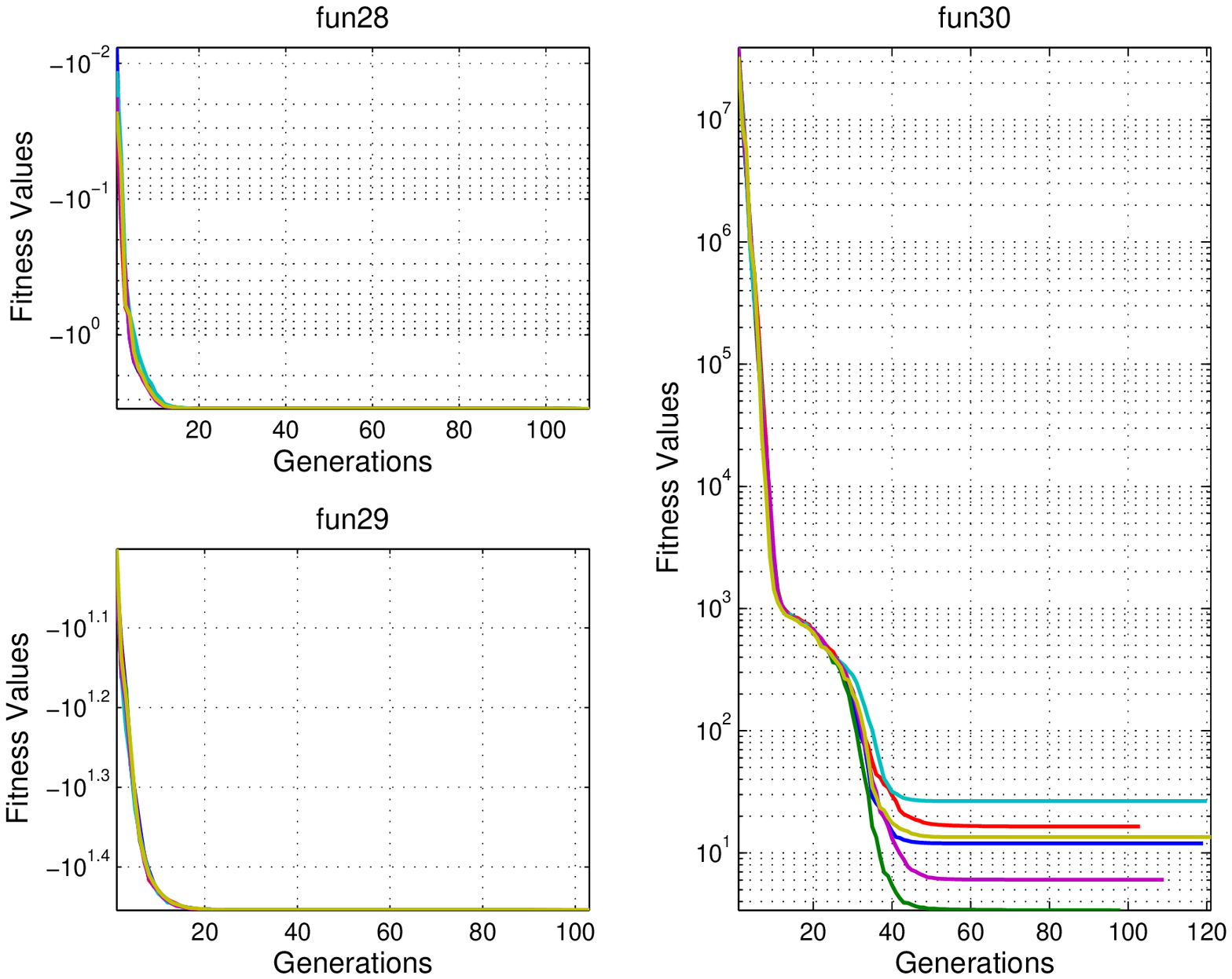}
  \end{figure}

%
\begin{table*}																																															 \centering

\caption{ Average, Standard Deviation  of the Best Fitness values for $f_{1}-f_{8}$.}		\label{at1}																																													 \small{
\begin{tabular*}{\textwidth}{@{\extracolsep{\fill}}l|l|r|r|r|r|r|r|r|r}																																															
\hline																																															
		Function	&		&	$f_1$	&	$f_2$	&	$f_3$	&	$f_4$	&	$f_5$	&	$f_6$	&	$f_7$	&	$f_8$	\\																										 \hline
		Dimension	&		&	30 	&	30 	&	30 	&	30 	&	30 	&	30 	&	30 	&	30 	\\																										 \hline
		Optimum	&		&	0.0000 	&	0.0000 	&	0.0000 	&	0.0000 	&	0.0000 	&	0.0000 	&	0.0000 	&	0.0000 	\\																										 \hline
		HdEA	&	average	&	0.0000 	&	0.0000 	&	16920.2300 	&	10.8802 	&	21.1276 	&	10.4615 	&	0.0000 	&	 0.0000 	\\																										
			&	std. dev.	&	0.0000 	&	0.0000 	&	2818.2200 	&	1.3212 	&	13.7111 	&	0.6028 	&	0.0000 	&	0.0000 	\\																																																			
	\hline	
RCGA	&	average	&	0.0000 	&	0.0000 	&	2811.5869 	&	91.8556 	&	71.1275 	&	8.8836 	&	219.8173 	&	1.6946 	 \\																										
-UNDX			&	std. dev.	&	0.0000 	&	0.0000 	&	1668.9500 	&	20.8748 	&	67.2881 	&	0.4382 	&	12.6701 	&	 0.1151 	\\																																																			
	\hline	CMA-ES	&	average	&	0.0000 	&	0.0000 	&	0.0000 	&	0.0000 	&	0.0000 	&	0.2303 	&	53.6481 	&	 0.0014 	\\																										
			&	std. dev.	&	0.0000 	&	0.0000 	&	0.0000 	&	0.0000 	&	0.0000 	&	0.0893 	&	14.1662 	&	0.0036 	\\																										
																									 \hline
		DE	&	average	&	0.0221 	&	0.3594 	&	26073.0400 	&	49.5082 	&	614.4588 	&	10.9846 	&	25.7105 	&	 0.9931 	\\																										
			&	std. dev.	&	0.0048 	&	0.0371 	&	3339.0600 	&	4.0114 	&	112.1748 	&	0.5641 	&	2.6500 	&	0.0301 	 \\																										
																													
	\hline	ODE	&	average	&	0.0000 	&	0.0953 	&	78.1691 	&	0.0111 	&	27.0344 	&	8.5199 	&	102.3277 	&	 0.0258 	\\																										
			&	std. dev.	&	0.0000 	&	0.0486 	&	45.1045 	&	0.0808 	&	0.7121 	&	0.4121 	&	37.9763 	&	0.0326 	\\																										
																												
	\hline	DEahcSPX	&	average	&	0.1075 	&	0.0322 	&	65.9908 	&	15.5882 	&	3160.5891 	&	8.7536 	&	40.8070 	&	 0.1613 	\\																										
			&	std. dev.	&	0.3990 	&	0.1423 	&	65.2220 	&	3.5293 	&	9387.6000 	&	0.5200 	&	26.7653 	&	0.2685 	\\																										
																													
	\hline	DPSO	&	average	&	3.3462 	&	8.8458 	&	1955.2153 	&	10.5390 	&	12789.4900 	&	9.9202 	&	 125.4958 	&	4.0567 	\\																										
			&	std. dev.	&	0.9949 	&	1.1698 	&	24.5853 	&	1.3743 	&	78.0953 	&	0.8192 	&	3.9027 	&	0.8773 	\\																										
																													
	\hline	SEPSO	&	average	&	2.8527 	&	8.8184 	&	2984.5912 	&	13.8670 	&	10464.0700 	&	10.6244 	&	 112.9555 	&	3.6995 	\\																										
			&	std. dev.	&	0.9324 	&	1.3980 	&	29.6756 	&	1.7487 	&	80.0175 	&	0.8755 	&	4.5789 	&	0.9511 	\\																										
																												
	\hline	EDA	&	average	&	3439.5320 	&	22.2520 	&	3749.1330 	&	21.1420 	&	30214.7330 	&	100.0690 	&	 188.3840 	&	30.5040 	\\																										
			&	std. dev.	&	1221.2100 	&	5.4338 	&	1294.7400 	&	5.9330 	&	12162.0800 	&	52.0020 	&	20.5550 	&	 10.9870 	\\																										
																												
	\hline	
HisEA	&	average	&	0.0000 	&	0.0000 	&	319.4241 	&	0.9769 	&	28.0095 	&	0.0002 	&	0.0800 	&	0.0031 	\\
	&	std.dev.	&	0.0000 	&	0.0000 	&	242.1950 	&	0.2552 	&	1.5109 	&	0.0004 	&	0.2425 	&	0.0042 	\\

\hline	

\end{tabular*}																																															 }
\end{table*}		
																																												
\begin{table*}																																															
\caption{ Average, Standard Deviation  of the Best Fitness values for $f_{9}-f_{16}$.}		\label{at2}																																														
\small{																																															
\begin{tabular*}{\textwidth}{@{\extracolsep{\fill}}l|l|r|r|r|r|r|r|r|r}  																																															
	\hline																																														
																																															
		function 	&		&	$f_{9}$	&	$f_{10}$	&	$f_{11}$	&	$f_{12}$	&	$f_{13}$	&	$f_{14}$	&	$f_{15}$	&	$f_{16}$	\\																										
	\hline	Dimension	&		&	30	&	30	&	2	&	2	&	2	&	30	&	30	&	30	\\																										
	\hline	Optimum	&		&	$-$	&	0.0000 	&	-1.0316 	&	0.3980 	&	3.0000 	&	0.0000 	&	0.0000 	&	0.0000 	\\																										
	\hline	HdEA	&	average	&	-1.37E+04	&	0.00E+00	&	-1.0316	&	4.01E-01	&	4.41E+00	&	0.00E+00	&	0.00E+00	&	 2.61E+02	\\																										
			&	std. dev.	&	2.54E+01	&	0.00E+00	&	1.00E-04	&	2.64E-02	&	4.08E+00	&	0.00E+00	&	0.00E+00	&	3.50E+01	\\																																																			
	\hline	RCGA	&	average	&	-5.94E+03	&	2.07E+01	&	-0.6587	&	4.60E-01	&	5.61E+01	&	6.92E+07	&	3.33E-01	&	 2.94E+02	\\																										
	-UNDX		&	std. dev.	&	4.87E+02	&	9.04E-02	&	3.12E-01	&	5.65E-02	&	2.96E+01	&	1.20E+07	&	9.54E-02	&	 3.69E+01	\\																										
																											
	\hline	CMA-ES	&	average	&	-5.40E+03	&	2.13E+01	&	-1.0235	&	3.98E-01	&	7.32E+00	&	4.14E+01	&	1.15E-02	&	 0.00E+00	\\																										
			&	std. dev.	&	9.56E+01	&	4.32E-01	&	8.16E-02	&	0.00E+00	&	1.66E+01	&	1.01E+02	&	1.15E-01	&	0.00E+00	\\																										
																												
	\hline	DE	&	average	&	-1.28E+04	&	1.83E+00	&	-0.6695	&	1.52E+00	&	1.48E+01	&	4.66E+03	&	3.20E-03	&	 2.74E+02	\\																										
			&	std. dev.	&	1.56E+02	&	3.32E-01	&	3.21E-01	&	1.35E+00	&	1.04E+01	&	9.26E+02	&	8.00E-04	&	3.00E+01	\\																										
																												
	\hline	ODE	&	average	&	-5.47E+03	&	9.90E-03	&	-1.0214	&	4.25E-01	&	3.52E+00	&	1.16E+00	&	1.50E-05	&	 2.43E+01	\\																										
			&	std. dev.	&	5.06E+02	&	1.17E-02	&	1.09E-02	&	2.77E-02	&	4.77E-01	&	1.25E+00	&	0.00E+00	&	8.07E+00	\\																										
																													
	\hline	DEahc-&	average	&	-1.02E+04	&	2.82E+00	&	-0.4882	&	6.68E+00	&	2.18E+01	&	3.82E+04	&	5.98E-02	&	 2.21E+00	\\																										
			SPX	&	std. dev.	&	6.92E+02	&	4.03E+00	&	3.29E+00	&	1.93E+01	&	8.50E+01	&	2.01E+05	&	1.70E-01	&	 3.85E+00	\\																										
																												
	\hline	DPSO	&	average	&	-5.18E+03	&	6.07E+00	&	-1.0229	&	1.44E+00	&	3.14E+00	&	8.26E+06	&	1.31E+01	&	 1.35E+02	\\																										
			&	std. dev.	&	2.57E+01	&	8.21E-01	&	1.17E-01	&	1.62E+00	&	6.47E-01	&	1.93E+03	&	1.66E+00	&	7.91E+00	\\																										
																												
	\hline	SEPSO	&	average	&	-7.70E+03	&	6.44E+00	&	-1.0252	&	4.10E-01	&	3.06E+00	&	6.28E+06	&	1.40E+01	&	 7.18E+01	\\																										
			&	std. dev.	&	2.73E+01	&	1.00E+00	&	1.04E-01	&	1.37E-01	&	2.93E-01	&	1.55E+03	&	1.45E+00	&	4.87E+00	\\																										
																											
	\hline	EDA	&	average	&	-4.67E+03	&	1.02E+01	&	-1.031	&	3.98E-01	&	3.00E+00	&	6.75E+06	&	7.42E+04	&	 6.97E+01	\\																										
			&	std. dev.	&	7.03E+02	&	1.29E+00	&	1.20E-03	&	0.00E+00	&	2.00E-06	&	4.13E+06	&	5.60E+04	&	2.91E+01	\\																										
																												
	\hline	
HisEA	&	average	&	-12558.8751 	&	0.0000 	&	-1.0316 	&	0.3979 	&	3.0000 	&	2.46E+05	&	0.0000 	&	0.3054 	\\
	&	std.dev.	&	29.1137 	&	0.0000 	&	0.0000 	&	0.0000 	&	0.0000 	&	1.91E+05	&	0.0000 	&	0.1539 	\\
																						
\hline																																															
\end{tabular*}																																															
}																																															
\end{table*}

\begin{table*}																																															
\caption{ Average, Standard Deviation  of the Best Fitness values for $f_{17}-f_{24}$.}		\label{at3}																																														
\small{																																															
\begin{tabular*}{\textwidth}{@{\extracolsep{\fill}}l|l|r|r|r|r|r|r|r|r}  																																															 \hline
		Function	&		&	$f_{17}$	&	$f_{18}$	&	$f_{19}$	&	$f_{20}$	&	$f_{21}$	&	$f_{22}$	&	$f_{23}$	&	$f_{24}$	\\																										 \hline
		Dimension	&		&	30 	&	30 	&	30 	&	30 	&	30 	&	30 	&	30 	&	30 \\																											 \hline
	Optimum	&		&	0.0000 	&	0.0000 	&	-29.0000  	&	0.0000  	&	$-$	&	0.0000  	&	-4930.0000 	&	$-$	\\																										
	\hline	
        HdEA	&	average	&	0.0004	&	4.8663	&	-24.9443	&	0.0000	&	-25.3678	&	0.1626	&	8025.425	&	-997867	\\																										
			&	std. dev.	&	0.0002	&	0.3451	&	0.9411	&	0.0000	&	0.572	&	0.4616	&	4773.2	&	0.0271	\\																										
	\hline	RCGA-	&	average	&	10.2837	&	7.61	&	-6.633	&	0.3566	&	-8.8451	&	152.9672	&	62837.03	&	-330	\\
		UNDX	&	std.                                       dev.	&	1.3909	&	0.5734	&	0.452	&	0.073	&	0.5691	&	17.6976	&	3012.7	&	 0.0000	\\
			
	\hline	CMA-ES	&	average	&	0.0025	&	13.7823	&	-0.9678	&	0.4493	&	-19.1834	&	0.023	&	-2428.19	&	-951	\\
			&	std.                                       dev.	&	0.0028	&	0.2792	&	0.732	&	0.258	&	1.8797	&	0.0472	&	0.0000	&	187570.7	\\
			
\hline		
DE	&	average	&	0.1641	&	5.3987	&	-18.8816	&	0.0027	&	-18.3183	&	60.0966	&	122598.2	&	-958473	\\
			&	std.                       dev.	&	0.0486	&	0.5198	&	0.556	&	0.0006	&	0.6445	&	10.9953	&	25422.81	&	6695.09	\\
			
	\hline	ODE	&	average	&	0.0299	&	0.0237	&	-27.8856	&	0.000027	&	-12.5543	&	26.0994	&	-4930	&	-610112	\\
			&	std.                               dev.	&	0.0099	&	0.1431	&	1.8404	&	0	&	1.2739	&	12.9456	&	1162.05	&	37311.6	\\
			
	\hline	DEahc-	&	average	&	0.0013	&	4.6963	&	-14.6745	&	0.1752	&	-12.9365	&	37.1675	&	1911.297	&	-996116	\\
		SPX	&	std.                           dev.	&	0.0078	&	1.3226	&	4.0927	&	0.1499	&	2.0401	&	17.6322	&	4085.18	&	 5578.9	\\
			
	\hline	DPSO	&	average	&	5.758	&	11.8132	&	-15.4114	&	0.6795	&	-10.3292	&	135.0221	&	26342.66	&	-342933	 \\
			&	std.                           dev.	&	1.3801	&	0.521	&	1.2455	&	0.4754	&	0.8904	&	6.6864	&	86.3998	&	217.3297	\\
			
	\hline	SEPSO	&	average	&	9.5052	&	12.0147	&	-16.9436	&	0.7684	&	-10.9954	&	152.5561	&	30572.09	&	-965431	 \\
			&	std.                           dev.	&	1.7957	&	0.532	&	1.2194	&	0.5383	&	0.9763	&	7.1358	&	100.8033	&	24.4007	\\
			
	\hline	EDA	&	average	&	12.235	&	12.309	&	-18.728	&	1.885	&	-9.361	&	10.527	&	141156.77	&	-286765.1	\\
			&	std.   dev.	&	3.110696	&	0.17794	&	4.185106	&	0.444738	&	0.75983	&	4.54527	&	65517.19	&	36881.35	\\
			
	\hline	
HisEA	&	average	&	0.0027 	&	2.6454 	&	-28.9299 	&	0.0000 	&	-25.9147 	&	0.0000 	&	2834.5739 	&	-984105.1432 	\\
	&	std.dev.	&	0.0018 	&	0.5052 	&	0.1942 	&	0.0000 	&	0.9994 	&	0.0000 	&	1627.9243 	&	3769.7065 	\\

\hline																					
\end{tabular*}																					
}																					
\end{table*}

\begin{table*}																																															
\caption{Average, Standard Deviation  of the Best Fitness values for $f_{25}-f_{30}$.}		\label{at4}																																														
\small{																																															
\begin{tabular*}{\textwidth}{@{\extracolsep{\fill}}l|r|r|r|r|r|r|r}  																																															
\hline																																															
																																															
	function 	&		&			$f_{25}$	&	$f_{26}$	&	$f_{27}$	&	$f_{28}$	&	$f_{29}$	&	$f_{30}$	\\																													
\hline	Dimension	&		&			30 	&	30 	&	30 	&	30 	&	30 	&	30 	\\																													
\hline	Optimum	&		&			0.9000 	&	0.0000 	&	$-$	&	-3.5000 	&	$-$ 	&	$-$	\\																													
\hline	HdEA	&	average	&			1.0004	&	1.2051	&	-2E+34	&	-1.521	&	-29.559	&	20.6012	\\																													
		&	std. dev.	&			0.0002	&	0.1365	&	3.60E+33	&	0.6748	&	0.0289	&	28.8726	\\																																																									
\hline	RCGA	&	average	&			6.9638	&	2.3127	&	-1.3E+20	&	-3.3678	&	-10.2002	&	7440.466	\\																													
	-UNDX	&	std. dev.	&			0.3607	&	0.1817	&	2.10E+20	&	0.0231	&	0.5695	&	2193.41	\\																													
																													
\hline	CMA-ES	&	average	&			8.142	&	1.1979	&	-1.1E+29	&	-2.6016	&	-19.1408	&	319.3721	\\																													
		&	std. dev.	&			5.7645	&	0.247	&	4.40E+29	&	1.5276	&	2.0299	&	102.4076	\\
		
\hline	DE	&	average	&			1.4523	&	3.6819	&	-9E+29	&	-2.1663	&	-24.8678	&	830.2062	\\
		&	std. dev.	&			0.076	&	0.2572	&	1.60E+30	&	0.1757	&	0.418	&	15.6812	\\
		
\hline	ODE	&	average	&			0.9107	&	0.4718	&	-1.2E+24	&	-3.5000	&	-14.7206	&	766.9481	\\
		&	std. dev.	&			0.0418	&	0.1056	&	7.10E+24	&	0.0000	&	1.0599	&	22.6609	\\
\hline
	DEahcSPX	&	average	&			2.4177	&	0.4953	&	-1.9E+24	&	-3.3078	&	-16.9775	&	9536.837	\\
		&	std. dev.	&			0.8642	&	0.1756	&	1.20E+25	&	0.3708	&	2.3975	&	38054.29	\\
		
\hline	DPSO	&	average	&			4.8472	&	2.9157	&	-2.7E+24	&	-1.8368	&	-13.6114	&	11716.56	\\
		&	std. dev.	&			0.8796	&	0.6078	&	4.10E+12	&	0.7575	&	0.965	&	77.5304	\\
		
\hline	SEPSO	&	average	&			3.3681	&	3.3948	&	-2.1E+25	&	-2.3102	&	-14.0837	&	11007.86	\\
		&	std. dev.	&			0.7988	&	0.7648	&	8.60E+12	&	0.8684	&	1.1034	&	81.4433	\\
		
\hline	EDA	&	average	&			7.629	&	5.186	&	-1.24E+22	&	-1.222	&	-10.781	&	985056.31	\\
		&	std. dev.	&			0.5443	&	1.0724	&	8.90E+22	&	0.2879	&	0.7085	&	769131.6	\\
		
\hline	
HisEA	&	average	&	1.0000 	&	0.1859 	&	-6.25E+34	&	-3.5000 	&	-28.4301 	&	13.5723 \\
	&	std.dev.	&	0.0000 	&	0.0351 	&	9.93E+30	&	0.0000 	&	0.0179 	&	11.8222 \\
\hline																		
\end{tabular*}																		
}																		
\end{table*}

\subsubsection{Test functions}
30 well-known real valued functions are used to evaluate the performance of HisEA in this paper.
The test functions, the numbers of dimensions, and the ranges of search are as follows.

\begin{align}
f_1(x) & = \sum_{i=1}^{n} x_{i}^{2}, & n &= 30,  x \in [-100,100]^n.
\end{align}
\begin{align}
f_2(x)  & = \sum_{i=1}^n |x_i|+ \prod_{i=1}^{n} |x_i|,   & n & =30,   x \in[-10,10]^n.
\end{align}
\begin{align}
f_3(x) & = \sum_{i=1}^{n} \left( \sum_{j=1}^{i} x_j \right)^2, & n &=30,    x   \in [-100,100]^n. \end{align}
\begin{align}
f_4(x) & = \max _{i \in [1,n]}|x_i|, & n &=30,    x  \in [-100,100]^n.
 \end{align}
\begin{align}
 f_5(x) & = \sum_{i=1}^{n-1}  \left[ 100(x_{i+1}-x_{i}^2)^2 + (x_i - 1)^2  \right],   \nonumber \\
        & n =30,     x   \in[-29, 31]^n.
  \end{align}
\begin{align}
f_6(x) & =\sum_{i=1}^{n} i x_{i}^4, & n &=30,   x  \in[-1.28,1.28]^n.
 \end{align}
\begin{align}
f_7(x) & =  \sum_{i=1}^n \left[ x_{i}^2 -10\cos(2\pi x_i) +10 \right], \nonumber \\
&  n  =30,   x \in [-5.12,5.12]^n.
\end{align}
\begin{align}
f_8(x) & = \frac{1}{4000} \sum_{i=1}^n x_i^2 - \prod_{i=1}^n \cos(\frac{x_i}{\sqrt{i}})+1, \nonumber \\
     & n=30,   x \in [-600,600]^n.
 \end{align}
\begin{align}
f_9(x) & =-\sum_{i=1}^n x_i \sin(\sqrt{|x_i|}), \nonumber \\
      & n  =30,   x  \in [-500,500]^n.
 \end{align}
\begin{align}
f_{10}(x) & =-20 \exp\left( -0.2 \sqrt{\frac{1}{n}\sum_{i=1}^n x_i^2} \right)     \nonumber \\
          & \quad - \exp\left( \frac{1}{n} \sum_{i=1}^n \cos(2\pi x_i)  \right) + 20 + e,  \nonumber \\
           & n  =30,   x   \in [-32,32]^n.
 \end{align}
\begin{align}
f_{11}(x)   = 4x_1^2-2.1x_1^4+\frac{1}{3}x_1^6+x_1x_2-4x_2^2+4x_2^4, \nonumber \\
  n  =2,  x  \in [-4.91017,5.0893] \times [-5.7126,4.2874].
 \end{align}
\begin{align}
f_{12}(x) & =\left( x_2 -\frac{5}{4\pi^2} x_1^2 + \frac{5}{\pi} x_1 - 6 \right)^2 \nonumber \\
   & \quad +10\left( 1-\frac{1}{8\pi} \right) \cos{x_1} + 10, \nonumber \\
& n  =2,   x  \in [-8.142,6.858]\times[-12.275,2.725].
\end{align}
\begin{align}
f_{13}(x) & =\left(  1+(x_1+x_2+1)^2 \right.\nonumber \\
        & \quad \cdot  \left. (19-14x_1+3x_1^2-14x_2+6x_1x_2+3x_2^2)  \right) \nonumber \\
   & \quad \times \left(  30+(2x_1-3x_2)^2 \right. \nonumber \\
  & \quad \cdot \left. (18-32x_1+12x_1^2+48x_2-36x_1x_2+27x_2^2 )\right),  \nonumber \\
    & n=2, x \in [-2,2]\times[-3,1].
\end{align}
\begin{align}
f_{14}(x) &=\sum_{i=1}^{n} 10^{\frac{6(i-1)}{n-1}}(x_i+100)^2,\nonumber \\
 & n=30, x \in [-100,100]^n.
\end{align}
\begin{align}
 f_{15}(x) & =g(x)+h(x), \text{where} \nonumber \\
g(x)& =\sin^2(\pi x_1) + \sum_{i=1}^{n-1} \left(x_i - 1\right)^2 \left(1+10\sin^2(\pi x_{i+1})\right), \nonumber \\
h(x) & =\left(x_n-1\right)^2\left(1+10\sin^2(2 \pi x_n)\right), \nonumber \\
 & n=30, x\in[-10,10]^n.
\end{align}
\begin{align}
f_{16}(x) &=\sum_{i=1}^{n} x_{i}^2 +\left( \sum_{i=1}^{n} 0.5 i x_i \right)^2 +\left( \sum_{i=1}^{n} 0.5 i x_i \right)^4, \nonumber \\
 & n=30, x \in [-5,10]^n.
\end{align}
\begin{align}
f_{17}(x) & = \sum_{i=1}^{n} \mid  x_i \sin(x_i) + 0.1 x_i   \mid,  \nonumber \\
 & n=30, x\in [-10,10]^n.
\end{align}
 \begin{align}
f_{18}(x) & = \sum_{i=1}^{n-1} \left( 0.5 + \frac{ \sin^{2}\sqrt{100x_i^2 +x_{i+1}^2} -0.5}{1+0.001\left( x_i^2-2x_ix_{i+1} + x_{i+1}^2 \right)^2}  \right), \nonumber \\
& n=30, x\in  [-100,100]^n.
\end{align}
\begin{align}
f_{19}(x) & = -\sum_{i=1}^{n-1} \left( \exp\left( \frac{-(x_i^2 + x_{i+1}^2 + 0.5 x_i x_{i+1})}{8}  \right)\right.  \nonumber \\
  & \quad \times \left. \cos\left( 4\sqrt{x_i^2 +x_{i+1}^2 + 0.5 x_i x_{i+1}}  \right) \right),  \nonumber \\
& n=30, x\in [-5,5]^n.
\end{align}
\begin{align}
f_{20}(x) & = 0.1n-\left( 0.1 \sum_{i=1}^{n} \cos(5\pi x_i) - \sum_{i=1}^{n} x_i^2 \right), \nonumber \\
          & n=30, x \in [-1,1]^n.
\end{align}
\begin{align}
f_{21}(x) & = - \sum_{i=1}^{n} \sin\left(y_i^2\right) \sin^{2m}\left( \frac{iy_i^2}{\pi} \right), \text{where} \nonumber \\
 y_i & =
 \left\{
   \begin{array}{ll}
      x_i \cos(\theta)-x_{i+1}\sin(\theta), &   i=1,3,5,\cdots <n; \\
      x_i \sin(\theta) + x_{i+1} \cos(\theta), &   i=2,4,6,\cdots<n;  \\
      x_i, &  i=n;
    \end{array}
  \right. \nonumber \\
 & m=10, \theta=\pi/6, n=30, x\in[0,\pi]^n.
\end{align}
\begin{align}
f_{22}(x) & = 0.1\left(h(x)+g(x)\right), \text{where} \nonumber \\
    h(x) & = \sin^2 (3\pi x_1) + (x_n -1)^2 (1+ 10\sin^2 (2\pi x_n)), \nonumber \\
    g(x) & = \sum_{i=1}^{n-1} (x_i - 1)^2 (1+ 10 \sin^2 (3\pi x_{i+1})),  \nonumber \\
& n=30, x \in [-5,5]^n.
\end{align}
\begin{align}
f_{23}(x) & = \sum_{i=1}^{n} (x_i - 1)^2 -\sum_{i=2}^{n} x_{i}x_{i-1}, \nonumber \\
 & n=30, x\in [-n^2, n^2]^n.
\end{align}
\begin{align}
f_{24}(x) & = \sum_{i=1}^{n} \left[ \ln^2 \left(x_i-2  \right) + \ln^2( 10-x_i) \right] - \left(\prod_{i=1}^{n} x_i\right)^{0.2}, \nonumber \\
 & n=30, x \in [2,10]^n.
\end{align}
\begin{align}
f_{25}(x) & = 1+\sum_{i=1}^n \sin^2(x_i) - 0.1\prod_{i=1}^{n} \exp(-x_i^2), \nonumber \\
 & n=30, x\in[-10,10]^n.
\end{align}
\begin{align}
f_{26}(x) & = 1- \cos\left( 2\pi \sqrt{\sum_{i=1}^{n}x_{i}^2}  \right) + 0.1 \sqrt{\sum_{i=1}^{n} x_{i}^2}, \nonumber \\
 & n=30, x\in[-100,100]^n.
\end{align}
\begin{align}
f_{27}(x) & = \prod_{i=1}^{n} \left( \sum_{j=1}^{5} j \cos\left( (j+1)x_i +j\right)  \right), \nonumber \\
 & n=30, x\in [-10,10]^n.
\end{align}
\begin{align}
f_{28}(x) & = -2.5\prod_{i=1}^n \sin\left( \frac{(x_i -30)\pi}{180} \right) \nonumber \\
 & \quad - \prod_{i=1}^{n} \sin\left( \frac{5(x_i- 30)\pi}{180} \right), \nonumber \\
 & n=30, x\in[0,180]^n.
\end{align}
\begin{align}
f_{29}(x) & = -\sum_{i=1}^{n} \left( \sin(x_i^2) \sin^{2m}\left( \frac{i x_{i}^2}{\pi} \right) \right), \nonumber \\
& m=10,  n=30, x\in [0,\pi]^n.
\end{align}
\begin{align}
f_{30}(x) & = \sum_{j=1}^{n} \sum_{i=1}^{n} \left( \frac{y_{i,j}}{4000} - \cos(y_{i,j}) +1 \right), \nonumber \\
  y_{i,j} & = 100 \left( x_j -x_i^2 \right)^2 + \left( 1-x_i \right)^2, \nonumber \\
   & n=30, x \in [-100, 100 ]^n.
\end{align}
$f_{11}$, $f_{12}$, $f_{13}$ are 2-d functions, and the dimensions of other functions are all set to be $30$ in this paper.

\subsubsection{Parameter setting}
HisEA has only four parameters, which are specified as the size of first generation $N_f=5,000$, the number of partition for each
dimension $N_p=300$, the number of good genetics to be selected for the pool of good genetics $N_g=5,000$, and the base number of new samples for each genetic operator $N_s=5,000$.

\subsubsection{Convergence of HisEA}
HisEA is an iterated algorithm, which may convergence after several loops. There are six paths plotted for each test function to describe the process of search and the convergence of algorithm.

As the population size of HisEA is much larger than those algorithms in \cite{ChowYuen2011} and the references therein, the convergence of HisEA for lower dimension problems is much quickly, see Fig. \ref{lowerd}.

$f_{11}$, $f_{12}$ and $f_{13}$ are two-dimensional functions, and the minimal values of the three test functions appear about the third or fourth run, where no more than $640,000$ individuals are evaluated.  And the algorithm stops within 21 runs for these three functions, such that the total number of evaluations is not larger than $3,360,000$  for each test functions. The extra number of evaluations are devoted to reduce the risk of trapping  by local minimum solutions.

The convergence of HisEA for 30 dimensional test functions is described in Fig \ref{higher}. As the scale of population for each run is much larger than their counterparts, the convergence of HisEA  is much quickly.

\subsubsection{Accuracy comparison with other algorithms}

\begin{table*}																																															 \centering
\caption{Numbers of fitness values with wrong order of magnitude among 30 test functions for each algorithm.}\label{at5}																																													 \small{
\begin{tabular*}{0.9 \linewidth}{@{\extracolsep{\fill}}c|c|c|c|c|c|c|c|c|c|c}																																															\hline																																															Algorithm & HdEA	&	RCGA-UNDX	&	CMA-ES	&	DE	&	ODE	&	Deahc-SPX	&	DPSO	&	SEPSO	&	EDA	&	HisEA	\\
\hline
Number & 6	&	17	&	9	&	13	&	8	&	11	&	14	&	14	&	20	&	4	\\
\hline	
\end{tabular*}																																															 }
\end{table*}

The accuracy of HisEA is compared with  9 benchmark EAs reported in \cite{ChowYuen2011},  and the results except HisEA are taken from \cite{ChowYuen2011}. The  results are reported in Table \ref{at1} to Table \ref{at4}, where HisEA is applied to each test function 50 times,  and the average value and standard deviation are reported for each function.

There are 6 new minimum values  founded in our investigations, including $f_{6}$, $f_{19}$,  $f_{22}$, $f_{26}$, $f_{27}$, $f_{30}$, which are not reached by the algorithms involved in this paper. And there are 10 test functions, where the optimal fitness values given by HisEA are similar to those results given by other EAs, including $f_{1}, $$f_{2}$, $f_{10}$, $f_{11}$, $f_{12}$, $f_{13}$, $f_{15}$, $f_{20}$, $f_{21}$, $f_{28}$. There are also 8 test functions, whose fitness values given by HisEA are closed to those reported optimal values, including $f_{7}$, $f_{8}$, $f_{9}$, $f_{16}$, $f_{17}$, $f_{24}$, $f_{25}$, $f_{29}$.

On the other hand,  there are 6 test functions, where HisEA can not find the optimal values successfully, including $f_{3}$, $f_{4}$, $f_{5}$, $f_{14}$, $f_{18}$, $f_{23}$. However, HisEA has the smallest number of fitness values which are different from the optimal solutions with respect to the order of magnitude among the algorithms considered in this paper, see Table \ref{at5}.

 According to the ``no free lunch theorem for search'' \cite{WolperMacReady1996}, it is very hard to require a specified evolutionary algorithm to over performance all of the algorithms existing in the literatures.  It is more possible to find some specified algorithms for a particular problem.  The simulations in this paper indicated that there is an alternative choice to
 design evolutionary algorithms with importance sampling method based on the estimated distributional information from the history of search process.

%
%
%
\section{Conclusions and discussions}\label{sec-4}
A new hybrid evolutionary algorithm is proposed in this paper, where the distribution of optimal solutions
is estimated from a set of evaluated solutions, which are updated from one generation to the next.
 The selected good intervals are used to estimate the distribution of optimal solutions, where the sampling probabilities for good intervals and good genetics are determined through a cross validation mechanism, which is not dependent on the fitness values.
 And crossover, mutation and other stochastic operators are designed with importance sampling method.  In order to maintain the healthy of individuals, purely random sampling methods and evolutionary algorithms are also included in the proposed hybrid evolutionary algorithm.

30 benchmark test functions are used to evaluate the performance of the proposed algorithm in this paper. It is found that HisEA outperforms all of the other algorithms considered in this paper, where HisEA has the smallest number of fitness values which are different from the optimal solutions with respect to the order of magnitude among the algorithms considered in this paper.

Possible directions for future work include applying other methods to construct the marginal and joint distributions of the optimal solutions, and other sampling methods to generate new crossover and mutation operators. The third direction is to construct more efficiently update method to estimate the distribution of optimal solutions, where only part of the evaluated solutions are involved, and the requirement of memory is relatively small.

\section*{Acknowledgment}

{Guanghui Huang is supported by the Fundamental Research Funds for the Central Universities of China under Grant CDJZR10 100 007.}

\begin{IEEEbiography}[{\includegraphics[width=1in,height=1.25in,clip,keepaspectratio]{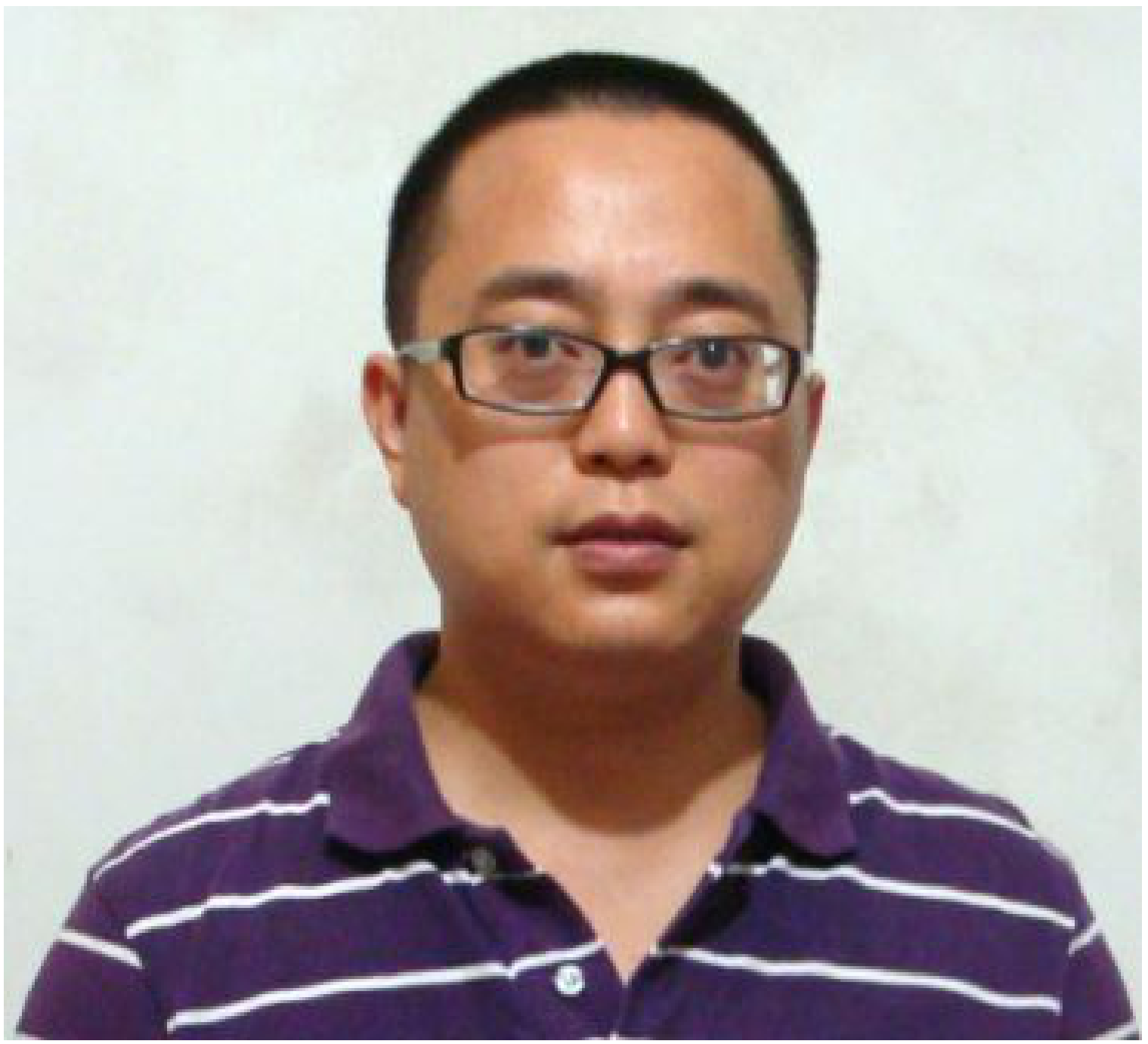}}]
{Guanghui Huang}
received the B.S. degree in applied mathematics from Southwest University, Chongqing, China, in 2001, and the Ph.D. in mathematics and statistics from Huazhong University of Science and Technology, Wuhan, China, in 2006.

He is currently an Associate Professor with the school of mathematics and statistics, Chongqing University, Chongqing, China. His current research interests include artificial intelligence algorithms, image processing, and Markov chain Monte Carlo simulations.
\end{IEEEbiography}

\begin{IEEEbiography}[{\includegraphics[width=1in,height=1.25in,clip,keepaspectratio]{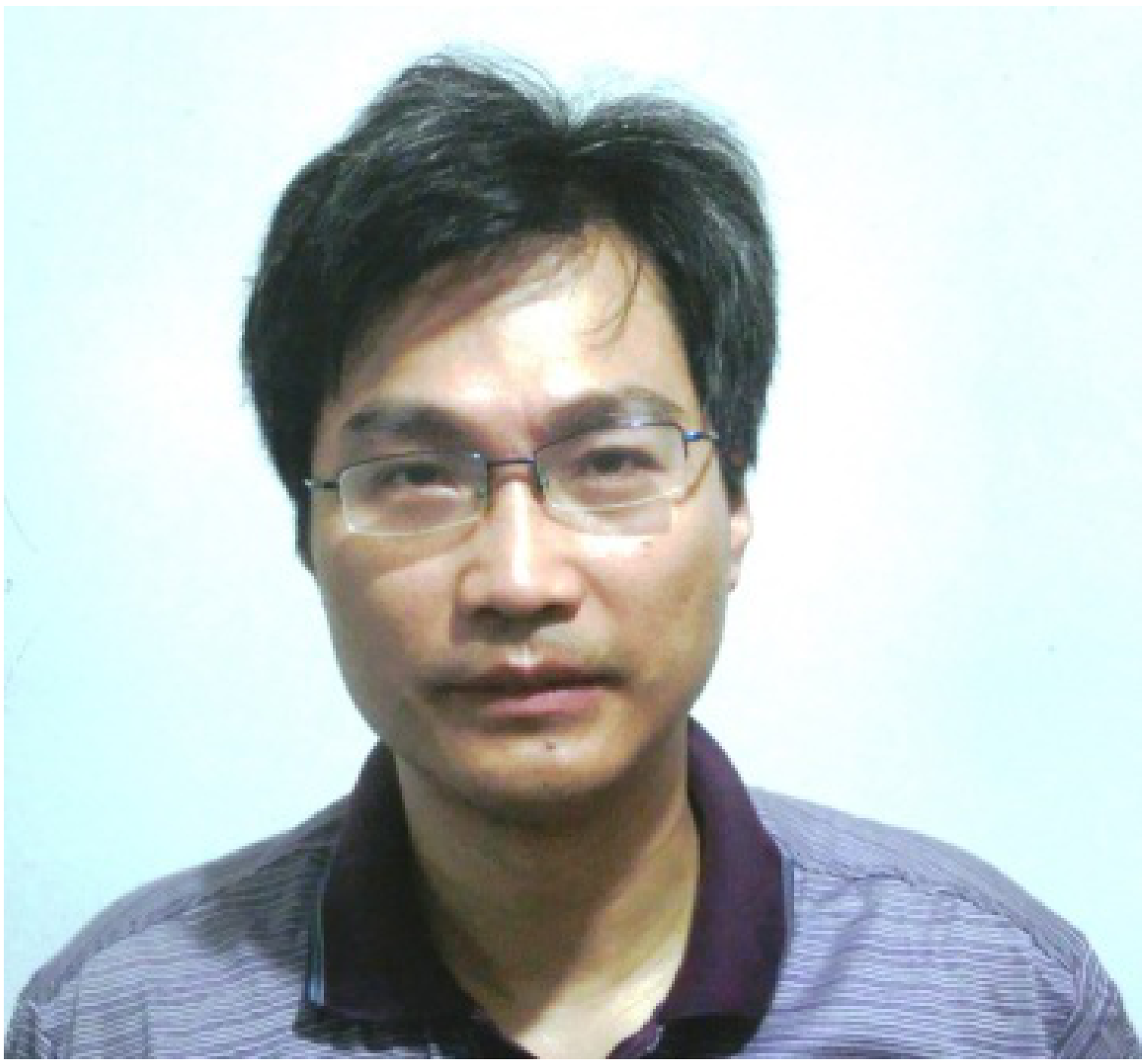}}]
{Zhifeng Pan}
received the B.S. degree in applied mathematics from Southwest University, Chongqing, China, in 1994, and the M.S. in mathematics and statistics from Chongqing University, Chongqing, China, in 2002.

He is currently an Assistant Professor with the school of mathematics and statistics, Chongqing University, Chongqing, China. His current research interests include artificial intelligence algorithms and Markov chain Monte Carlo simulations.
\end{IEEEbiography}




\end{document}